%% file: main.tex
\documentclass[10pt,twocolumn,letterpaper]{article}

\usepackage[pagenumbers]{cvpr} % To force page numbers, e.g. for an arXiv version

\input{preamble}

\definecolor{cvprblue}{rgb}{0.21,0.49,0.74}
\usepackage[pagebackref,breaklinks,colorlinks,allcolors=cvprblue]{hyperref}

\def\paperID{} % *** Enter the Paper ID here
\def\confName{CVPR}
\def\confYear{2026}

\title{GS4: Generalizable Sparse Splatting Semantic SLAM}

\author{
\href{https://mingqij.github.io/}{\color{Black}{Mingqi Jiang}}, 
\href{https://chkim403.github.io}{\color{Black}{Chanho Kim}},
\href{https://chenziwe.com}{\color{Black}{Chen Ziwen}}, 
\href{https://web.engr.oregonstate.edu/~lif/}{\color{Black}{Li Fuxin}}\\
Collaborative Robotics and Intelligent Systems (CoRIS) Institute\\
Oregon State University\\
{\tt\small   \{jiangmi, kimchanh, chenziw, lif\}@oregonstate.edu}\\
\small Project page: \url{https://mingqij.github.io/projects/gs4/}
}
\begin{document}
\maketitle
\input{sec/0_abstract}    
\input{sec/1_intro}
\input{sec/2_related_work}
\input{sec/3_method}

\input{sec/4_experiment}
\input{sec/5_conclusion}
{
    \small
    \bibliographystyle{ieeenat_fullname}
    \bibliography{main}
}

\input{sec/X_suppl}

\end{document}

%% file: preamble.tex
\usepackage{times}
\usepackage{epsfig}
\usepackage{graphicx}
\usepackage{amsmath}
\usepackage{amssymb}
\usepackage{amsfonts}
\usepackage{bbding}
\usepackage{colortbl}
\usepackage{arydshln}
\usepackage{booktabs}
\usepackage{multirow}

\newcommand{\redx}{{\color{red}\XSolidBrush}}
\colorlet{colorFst}{green!40}       %
\colorlet{colorSnd}{blue!30}      %
\colorlet{colorThi}{yellow!30}      %
\newcommand{\fs}{\cellcolor{colorFst}\bf}   %
\newcommand{\nd}{\cellcolor{colorSnd}}      %

\usepackage{subcaption}

\makeatletter
\newcommand{\cdashmidrule}[1]{%
  \noalign{\vskip\aboverulesep}
  \cdashline{#1}
  \noalign{\vskip\belowrulesep}}
\makeatother

%% file: sec/0_abstract.tex
\begin{abstract}
% Traditional SLAM algorithms are excellent at camera tracking but might generate lower resolution and incomplete 3D maps. Recently, Gaussian Splatting (GS) approaches have emerged as an option for SLAM with accurate, dense 3D map building. However, existing GS-based SLAM methods rely on per-scene optimization which is time-consuming and does not generalize to diverse scenes well. In this work, we introduce the first generalizable GS-based semantic SLAM algorithm that incrementally builds and updates a 3D scene representation from an RGB-D video stream using a learned deep network. Our approach starts from an RGB-D image recognition backbone to predict the Gaussian parameters from every downsampled and backprojected image location. Additionally, we seamlessly integrate 3D semantic segmentation into our GS framework, bridging 3D mapping and recognition through a shared  backbone. To correct localization drifting and floaters, we propose to optimize the GS for only few  iterations following global localization. We demonstrate state-of-the-art semantic SLAM performance on the real-world benchmark ScanNet with an order of magnitude fewer Gaussians compared to other recent GS-based methods, and showcase our model’s generalization capability through zero-shot transfer to the NYUv2 and TUM RGB-D datasets.
Traditional SLAM algorithms excel at camera tracking, but typically produce incomplete and low-resolution maps that are not tightly integrated with semantics prediction. Recent work integrates Gaussian Splatting (GS) into SLAM to enable dense, photorealistic 3D mapping, yet existing GS-based SLAM methods require per-scene optimization that is slow and consumes an excessive number of Gaussians. We present GS4, the first \textit{generalizable} GS-based semantic SLAM system. Compared with prior approaches, GS4 runs 10× faster, uses 10× fewer Gaussians, and achieves state-of-the-art performance across color, depth, semantic mapping and camera tracking. From an RGB-D video stream, GS4 incrementally builds and updates a set of 3D Gaussians using a feed-forward network. First, the Gaussian Prediction Model estimates a sparse set of Gaussian parameters from input frame, which integrates both color and semantic prediction with the same backbone. Then, the Gaussian Refinement Network merges new Gaussians with the existing set while avoiding redundancy. 
Finally, when significant pose changes are detected, we perform only 1–5 iterations of joint Gaussian–pose optimization to correct drift, remove floaters, and further improve tracking accuracy.
Experiments on the real-world ScanNet and ScanNet++ benchmarks demonstrate state-of-the-art semantic SLAM performance, with strong generalization capability shown through zero-shot transfer to the NYUv2 and TUM RGB-D datasets.
\end{abstract}

%% file: sec/1_intro.tex
\vspace{-0.15in}
\section{Introduction}
\label{sec:intro}

% Semantic Visual Simultaneous Localization and Mapping (SLAM) is a long-standing challenge in computer vision that aims to reconstruct a dense 3D semantic map of an environment while simultaneously estimating camera poses from an input video. These dense semantic maps have widespread applications in autonomous driving, AR/VR and robotics, enabling robots to understand and interact with their surroundings more effectively by providing rich 3D spatial and semantic information. %Previous SLAM works have utilized various 3D scene representations, such as feature point clouds~\cite{7946260}, voxel grids~\cite{6162880}, or neural scene representations~\cite{Sandström2023ICCV, Zhu2022CVPR}.

Simultaneous Localization and Mapping (SLAM) is a long-standing challenge in computer vision, aiming to reconstruct a 3D map of an environment while simultaneously estimating camera poses from a video stream. Semantic visual SLAM extends this goal by producing dense maps enriched with semantic labels, enabling applications in autonomous driving, AR/VR, and robotics. By combining geometric reconstruction with object-level understanding, semantic SLAM provides rich 3D spatial and semantic information that allows robots and other systems to navigate and interact with their surroundings more effectively.

Traditional visual SLAM systems consist of several independent components, including keypoint detection, feature matching, and bundle adjustment \citep{orb-slam,orbslam2,orb-slam3}.
% \todo{cite a few slams with voxel representations here}
Their scene representations are typically low-resolution voxels, which limit geometric detail. Thus, although these systems generally provide accurate camera localization, they struggle to generate dense, high-quality 3D maps, which are required for robotics applications such as mobile manipulation. Recent advances in differentiable rendering \citep{mildenhall2020nerf,gaussiansplatting} introduces new options for scene representation in visual SLAM. For example, neural scene representations such as Neural Radiance Fields (NeRF) \citep{mildenhall2020nerf} have been successfully adopted in SLAM frameworks \citep{imap,nice-slam,eslam}; however, NeRF requires hours of per-scene optimization, making it computationally expensive and forcing a trade-off between reconstruction quality and training cost.

\begin{figure}[t]
\centering
\includegraphics[width=0.8\linewidth]{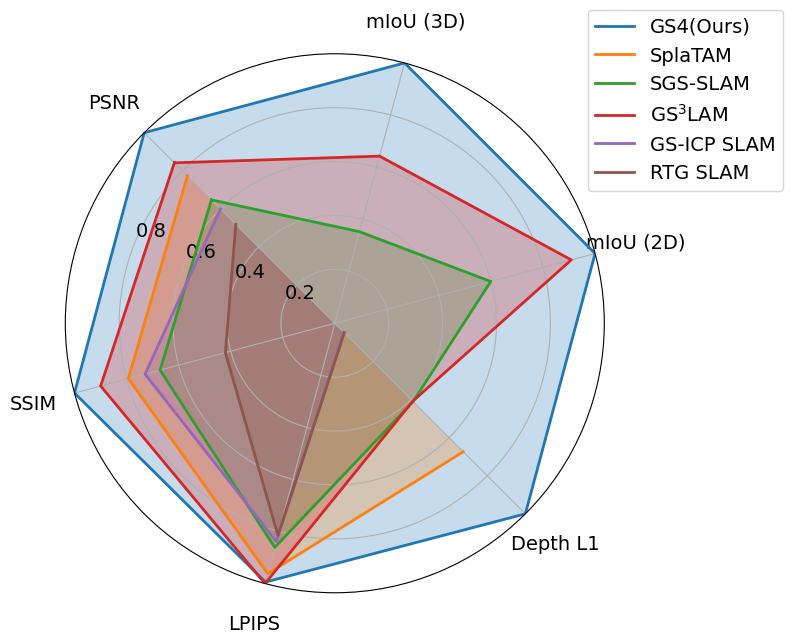}
\caption{\small A radar chart comparing rendering and 3d semantic metrics. We normalize each metric independently, values closer to the outer edge indicate better performance.}
\label{fig_radar}
\vskip -0.2in 
\end{figure}

\begin{figure}[t]
\centering
\includegraphics[width=0.8\linewidth]{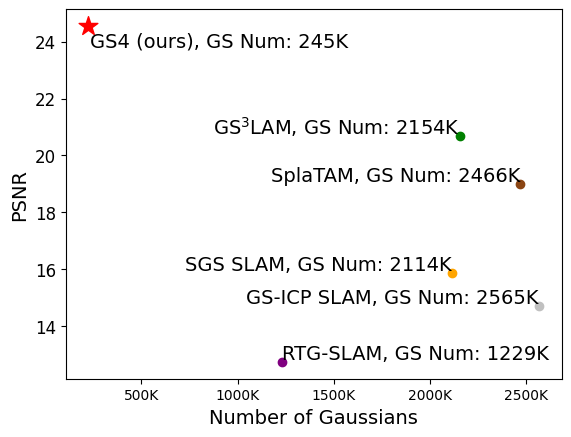}
\caption{\small Comparison of PSNR with respect to number of Gaussians across Gaussian Splatting SLAM algorithms (over an average of $2,680$ frames in the 6 testing scenes of ScanNet). Our method achieves state-of-the-art performance with much fewer Gaussians. GS Num represents the number of 3D Gaussians  in the scene after mapping is complete.}
\label{fig_psnrfps}
\vskip -0.2in 
\end{figure}

% \begin{figure}[t]
% \centering
% \includegraphics[width=0.7\linewidth]{figures/psnr_gsnum.png}
% \vskip -0.18in 
% \caption{\small A comparison of PSNR with respect to number of Gaussians used to track videos with an average of 2,680 frames across Gaussian Splatting SLAM algorithms. Our method achieves state-of-the-art performance in rendering evaluation with much fewer Gaussians. GS Num represents the number of 3D Gaussians  in the scene after mapping is complete. Reported values are averaged over the 6 testing scenes on ScanNet.
% }
% \label{fig_psnrfps}
% \vskip -0.2in 
% \end{figure}

% \begin{figure}[t]
% \centering
% \includegraphics[width=0.7\linewidth]{figures/radar.png}
% \vskip -0.18in 
% \caption{\small A radar chart comparing rendering metrics. We normalize each metric independently: for metrics where larger values are better, we divide by the maximum; for metrics where smaller values are better, we divide the minimum by the value. Values closer to the outer edge indicate better performance.}
% \label{fig_psnrfps}
% \vskip -0.2in 
% \end{figure}

Recently, 3D Gaussians have emerged as a powerful 3D scene representation, offering fast, differentiable, and high-quality rendering capabilities~\citep{kerbl3Dgaussians}. Leveraging these advantages, Gaussian-based representations have proven highly effective for SLAM systems~\citep{keetha2024splatam,Matsuki:Murai:etal:CVPR2024}. However, existing approaches still rely on test-time, gradient-based optimization to estimate 3D Gaussians for each scene independently, which is computationally expensive and unsuitable for real-time applications. In addition, these methods depend on heuristic Gaussian densification and pruning strategies~\citep{kerbl3Dgaussians}, often producing overly dense representations that fail to scale to large, real-world environments.

%While NeRF-based and GS-based SLAM approaches have demonstrated impressive 3D mapping results, integrating semantic segmentation remains rather cumbersome. These methods rely on per-scene optimization, requiring either running separate, independent semantic segmentation algorithms for input images first [TODO: cite] or ground truth segmentation masks [TODO: cite], which are then used as additional rendering targets. This dependence either ties the system’s performance to the accuracy of external segmentation algorithms or makes it impractical for real-world robotic applications, where ground truth segmentation masks are often unavailable. Additionally, it further complicates these SLAM systems.

In this paper, we propose \textbf{GS4} (\textbf{G}eneralizable \textbf{S}parse \textbf{S}platting \textbf{S}emantic \textbf{S}LAM), the first generalizable Gaussian-splatting–based SLAM system, which directly predicts 3D semantic Gaussians using a learned feed-forward network, eliminating the need for expensive per-scene optimization. By integrating an image recognition backbone, GS4 jointly reconstructs geometry, color, and semantic labels of the environment without relying on any external semantic-segmentation modules.

GS4 begins with the Gaussian Prediction Model that infers a sparse set of 3D semantic Gaussians from each incoming RGB-D frame in a feed-forward manner.
Next, the Gaussian Refinement Network integrates these newly predicted Gaussians with the evolving 3D map, replacing the handcrafted heuristics traditionally used for Gaussian densification and pruning. This learned refinement strategy yields a compact representation with an order-of-magnitude fewer Gaussians than competing methods.
Finally, after the global localization (bundle adjustment) step from the camera tracking module updates camera poses and Gaussian locations, we perform a lightweight few-iteration (only 1$\sim$5) optimization of Gaussian parameters to preserve rendering fidelity and mitigate the “floater” artifacts common in feed-forward GS approaches.

We demonstrate that GS4 achieves state-of-the-art performance across all key metrics in localization, mapping, and segmentation on the real-world benchmark ScanNet (Fig.~\ref{fig_radar}), while using only $\sim 10\%$ of Gaussians compared to prior GS SLAM methods (Fig.~\ref{fig_psnrfps}). Furthermore, we highlight the generalization capability of our system via zero-shot transfer to the NYUv2 and TUM RGB-D datasets, which, to the best of our knowledge, is the first demonstration of \textbf{zero-shot semantic SLAM} generalization in a modern neural SLAM system.

In summary, our contributions are as follows:
\begin{itemize}
\vspace{-0.03in}
\item We propose GS4, the first generalizable Gaussian splatting semantic SLAM approach on RGB-D sequences. Results showed that GS4 obtains state-of-the-art on real ScanNet and ScanNet++ scenes, and also zero-shot generalizes to the real NYUv2 and TUM RGB-D datasets without any fine-tuning.
\vspace{-0.01in}
\item Our proposed Gaussian refinement network effectively merges Gaussians from different frames into a 3D representation, while significantly reducing the number of Gaussians required to represent a scene to only $10\% - 25\%$ of prior work.%, with only $\sim 10\%$ of the Gaussians compared to prior work, effectively merging Gaussians predicted from different images into a consistent 3D representation.
\vspace{-0.01in}
\item Our proposed few-iteration joint Gaussian-pose optimization significantly improves reconstruction quality with a small additional computational cost, while also slightly improving tracking accuracy when high-quality ground-truth RGB-D data are available.
\end{itemize}
%Our experimental results demonstrate the feasibility of this method, showing a significant speed advantage, being \textbf{7x} faster than the existing method, SplaTAM~\cite{keetha2024splatam}.

%% file: sec/2_related_work.tex
\section{Related Work}
\label{sec:related_work}
\vspace{-0.04in}
\textbf{Traditional SLAM:}
Early visual SLAM methods \citep{orb-slam} demonstrated robust localization through effective keypoint detection and matching, which resulted in sparse 3D reconstructions. While these approaches provided reliable localization, the sparse nature of the reconstructed maps limited their utility in applications requiring detailed 3D maps. To address this issue, dense visual SLAM~\citep{kerl2013dense, czarnowski2020deepfactors} focused on constructing detailed maps to support applications like augmented reality (AR) and robotics. Prior methods \citep{canelhas2013sdf, dai2017bundlefusion, kinectfusion, bylow2013real, whelan2013robust, prisacariu2017infinitam} employ  representations based on Signed Distance Fields (SDF), %and Truncated Signed Distance Fields (TSDF), 
rather than relying on sparse representations such as point clouds or grids. However, these approaches often suffer from over-smoothed reconstruction, failing to capture fine details crucial for certain tasks.

\noindent\textbf{NeRF-based SLAM:} Neural Radiance Fields (NeRF)~\citep{mildenhall2020nerf} gained popularity as a 3D scene representation due to its ability to generate accurate and dense reconstructions. NeRF employs Multi-Layer Perceptron (MLP) to encode scene information and performs volume rendering by querying opacity and color along pixel rays. 
% ~\cite{mildenhall2021nerf} has emerged NeRF technique, enhances spatial modeling capabilities. 
Methods such as iMAP~\citep{imap}, NICE-SLAM~\citep{nice-slam}, and ESLAM~\citep{eslam} incorporate this implicit scene representation into SLAM, leveraging NeRF's high-fidelity reconstructions to improve both localization and mapping. DNS-SLAM~\citep{li2023dnsslamdenseneural} further incorporates semantic information into the framework. However, the volumentric rendering process in NeRF is costly, often requiring trade-offs such as limiting the number of pixels during rendering, These trade-offs, while improving efficiency, may compromise the system's accuracy in both localization and mapping. 

%Point-SLAM~\cite{pointslam}

\noindent\textbf{GS-based SLAM:} 
3D Gaussian Splatting (3DGS)~\citep{kerbl3Dgaussians} employs splatting rasterization instead of ray marching. This approach iterates over 3D Gaussian primitives rather than marching along rays%. %This approach leverages the inherent sparsity of 3D scenes, 
, resulting in a more expressive and efficient representation capable of capturing high-fidelity 3D scenes with significantly faster rendering speed. %Due to these desirable properties, recent SLAM methods have adopted 3D Gaussians as a scene representation~\cite{keselman2022approximatedifferentiablerenderingalgebraic, wang2022voge} and have extended these methods to dynamic scenes with dense 6-DOF motion~\cite{luiten2023dynamic, wu2024cvpr, yang2023deformable3dgs, yang2023gs4d}.
%With the efficiency and rapid rasterization capabilities of 3D Gaussian Splatting, 
Hence, GS-based SLAM systems achieve improved accuracy and speed in dense scene reconstruction. SplaTAM~\citep{splatam} introduces silhouette-guided rendering to support structured map expansion, enabling efficient dense visual SLAM. Gaussian Splatting SLAM~\citep{Matsuki:Murai:etal:CVPR2024} integrates novel Gaussian insertion and pruning strategies, while GS-ICP SLAM~\citep{ha2024rgbdgsicpslam} and RTG-SLAM~\citep{peng2024rtgslam} combine ICP with 3DGS to achieve both higher speed and superior map quality. Expanding upon these advancements, SGS-SLAM~\citep{li2024sgs}, OVO-SLAM~\citep{martins2024ovo},  SemGauss-SLAM~\citep{zhu2025semgaussslamdensesemanticgaussian} and GS$^3$LAM~\citep{li2024gs3lam} extend 3D Gaussian representations to include semantic scene understanding. However, existing GS-based SLAM methods employ per-scene optimization, requiring iterative refinement of Gaussians initialized from keyframes through rendering supervision. As a result, they all rely on additional segmentation models to predict semantic labels for each image, creating computational overhead.

\noindent\textbf{Feed-forward Models:}
Recently, MV-DUSt3R+~\citep{tang2024mv} and VGGT~\citep{wang2025vggt} have demonstrated multi-view inference by extending the DUSt3R~\citep{dust3r_cvpr24} architecture for multi-view reconstruction. Meanwhile, several works have introduced feed-forward approaches for scene-level 3DGS reconstruction using generalizable models~\citep{charatan23pixelsplat,chen2024mvsplat,liu2025mvsgaussian,gslrm2024,jiang2025anysplat,xu2024depthsplat}. In particular, pixelSplat~\citep{charatan23pixelsplat} predicts 3D Gaussians directly from image features, while DepthSplat~\citep{xu2024depthsplat} connects Gaussian splatting and depth estimation, studying their interaction to enable feed-forward 3D Gaussian reconstruction from multi-view images. AnySplat~\citep{jiang2025anysplat} further predicts both 3D Gaussian primitives and camera intrinsics/extrinsics for uncalibrated multi-view inputs. However, to the best of our knowledge, these feed-forward models have so far been applied only to a relatively small number of input images and have not been scaled to the thousands of frames typical in SLAM, nor have they been incorporated into GS-based semantic SLAM systems operating over such long sequences.

%% file: sec/3_method.tex
\begin{figure*}[htb]
\centering
\includegraphics[width=0.99\linewidth]{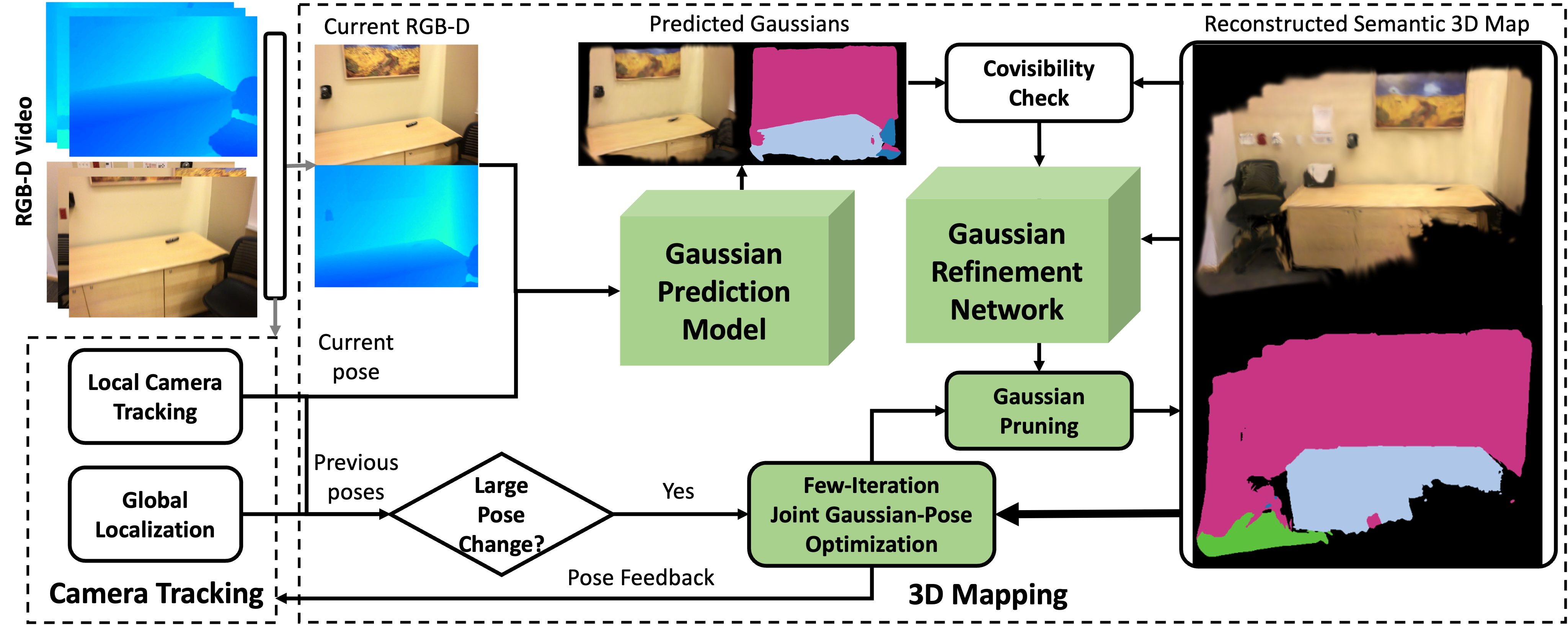}
% \vskip -0.15in 
\caption{\small \textbf{Overview of the SLAM System.} At each timestep, the system receives an RGB-D frame as input. The tracking system performs local camera tracking and global localization to determine the current frame's pose and correct previous pose errors. Our 3D mapping process comprises three main components: \textbf{1) Gaussian Prediction (Sec~\ref{sec:gs_prediction}):} Utilizing the current frame's RGB-D data,
% and the Plücker rays, 
%—computed from the frame's pose and camera intrinsics—
the Gaussian Prediction Model estimates the  parameters and semantic labels for all Gaussians in the current frame; \textbf{2) Gaussian Refinement (Sec~\ref{sec:gs_refinement}):} Both newly added Gaussians and those in the existing semantic 3D map are refined using the Gaussian Refinement Network to ensure that the combined set of Gaussians accurately represents the scene.  A covisibility check ensures that only non-overlapping Gaussians are integrated into the existing 3D map. Post-refinement, the transparent Gaussians are pruned; 
\textbf{3) Few-Iteration Joint Gaussian–Pose Optimization (Sec.~\ref{sec:gs_oneiter_opt})}: If significant pose corrections are detected, we perform a few iterations of joint Gaussian–pose optimization to update the Gaussians in the 3D map and further refine the poses; the refined poses are then fed back into the tracking system. This ensures consistency of the 3D map with the revised camera trajectories and further improves pose accuracy. (Best viewed in color.)
}
\label{fig_pipline}
\vskip -0.15in 
\end{figure*}
\vspace{-0.04in}
\section{Methods}
\vspace{-0.04in}
In this section, we describe our proposed SLAM approach. We first provide a brief overview of Gaussian Splatting, then detail our Gaussian prediction network and Gaussian refinement network. Finally, we explain how these networks are utilized within the entire SLAM system.%, including one-iteration Guassian optimization. An overview diagram of the proposed SLAM approach is presented in Fig.~\ref{fig_pipline}. %Given input RGB-D frames, we perform dense semantic mapping and real-time tracking.
\subsection{Gaussian Splatting}
We represent a 3D map using a set of anisotropic 3D Gaussians. Each Gaussian $G_i$ is characterized by RGB color $c_i \in \mathbb{R}^3$, center position $\mu_i \in \mathbb{R}^3$,
scale $s_i \in \mathbb{R}^{3}$, quaternion $r_i \in \mathbb{R}^{4}$,
% 3D covariance matrix $\mathbf{\Sigma_i} \in \mathbb{R}^{3 \times 3}$, 
opacity $ o_i \in \mathbb{R}$ and semantic class vector $v_i^{\text{class}} \in \mathbb{R}^N$, where N is the number of classes. 
% The 3D covariance matrix is parameterized as follows:
% \begin{equation}
%     \mathbf{\Sigma} = \mathbf{RSS}^{T}\mathbf{R}^{T},
%     \label{eq:covariance}
% \end{equation}
% where $\mathbf{S} \in \mathbb{R}^{3}$ is a 3D scale vector and $\mathbf{R} \in \mathbb{R}^{3\times3}$ is the rotation matrix, built using the quaternion representation.
% Given the viewing transformation $ \bf W$ and 3D covariance matrix $\mathbf{\Sigma}$, the 3D Gaussians are projected into the 2D image space for rendering using:
% \begin{equation}
%     \mathbf{\Sigma^{\prime}} = \mathbf{JW}^{-1}\mathbf{\Sigma W}^{-T}\mathbf{J}^{T},
%     \label{eq:covariance2d}
% \end{equation}
% where $\mathbf{J}$ is the Jacobian of the affine approximation of the projective function. After projecting 3D Gaussians to the image plane,  t

The rendering process is defined as:
\begin{equation*}
        Q_p  = \sum_{i \in N}q_i \alpha_i \prod_{j=1}^{i-1} (1-\alpha_j),
\end{equation*}
where $Q_p$ is a quantity of a pixel $p$ to be rendered, which can be color, depth or semantic label, and $q_i$ is that quantity of the $i$-th 3D Gaussian, while $\alpha_i$ is its visibility, computed from opacity and covariance parameters (determined by rotation and scale).
% \begin{equation*}
%     \alpha_i = o_i \text{exp}\big(-\frac{1}{2}\Delta_{i}^T \Sigma_{i}'^{-1} \Delta_{i}\big)
% \end{equation*}
% defines the visibility of the $i$-th splatted Gaussian. Here, $\Delta_{i} \in \mathbb{R}^{2}$ denotes the offset between the pixel coordinates and the 2D mean of the $i$-th Gaussian. 
%Similarly, the depth is rendered by:
%\begin{equation}
 %       D_p  = \sum_{i \in N}d_i \alpha_i \prod_{j=1}^{i-1} (1-\alpha_j),
%\end{equation}
%where $d_i$ denotes the depth of the center of the $i$-th 3D Gaussian.
Following \cite{keetha2024splatam}, We also render a silhouette image to determine visibility $
        S_p  = \sum_{i \in N} \alpha_i \prod_{j=1}^{i-1} (1-\alpha_j)$.
%Finally, the semantic image is rendered by:
%\begin{equation}
%        \text{class}_p  = max (\sum_{i \in N}v_i^{\text{class}} \alpha_i \prod_{j=1}^{i-1} (1-\alpha_j)),
%\end{equation}
%where $\text{class}_p$ denotes  the predicted class label of pixel $p$, while $v_i^{\text{class}}$ represents the semantic class vector of the $i$-th 3D Gaussian.

\subsection{Gaussian Prediction and Refinement}
Our proposed Gaussian prediction network (Fig.~\ref{fig_pipline}) takes RGB-D images as input and predicts 3D Gaussian parameters. Importantly, the backbone generates features that can predict semantic labels (e.g. trained from 2D segmentation tasks), enabling the rendering of photometric, geometric, and semantic views. Next, the Gaussian refinement network processes Gaussians predicted from a new frame and learns to merge them with the 3D scene representation computed from prior frames.%, constructing a consistent 3D scene representation.

\subsubsection{Backbone for Gaussian Prediction}
\label{sec:gs_prediction}
We train a transformer model to regress 3D GS parameters from an image with a known camera pose (from tracking, described in Sec.~\ref{sec:tracking}), while simultaneously assigning semantic labels to these 3D Gaussians. 
% We start with a 2D backbone, AutoFocusFormer~\cite{autofocusformer}, a local-attention transformer that adaptively downsamples by focusing on highly textured and salient regions, pre-trained for COCO instance segmentation using Mask2Former~\cite{cheng2021mask2former}. Following ODIN~\cite{jain2024odin}, we add 3D cross-view attention layers after the multi-scale attention stage. To further improve efficiency, we eliminate voxel-based operations and replace ODIN’s 3D interpolation step in these layers with the AFF method.
We start with a pre-trained 2D image segmentation model such as Mask2Former~\citep{cheng2021mask2former} or AutoFocusFormer~\citep{autofocusformer}, which encodes an image into encoder tokens $f_{enc}^l$ and decoder tokens $f_{dec}^l$ (from their image decoder) at several progressively downsampled levels $l = 1, \ldots, L$, with $L=4$ usually. %ODIN~\cite{jain2024odin}, a model that starts from multi-view RGB-D images with a transformer architecture that alternates between 2D within-view and 3D cross-view attention for information fusion over multiple views. %We replace the original 2D layers in ODIN with AutoFocusFormer (AFF) ~\cite{autofocusformer} which builds irregular 2D point cloud features rather than 2D grid features. AFF is pre-trained for COCO instance segmentation. 
We concatenate an RGB image $I \in R^{H\times W \times 3}$ and a depth image $D \in R^{H\times W \times 1}$
% , and Plücker coordinates $P \in R^{H\times W \times 6}$ \cite{ziwen2024llrm, gslrm2024}, 
resulting in a 4-channel feature map that is fed into the model:
\begin{equation*}
        \{f_{\text{enc}_i}^l, f_{\text{dec}_i}^l \}_{i=1:N_{\text{token}}}  = \text{Backbone}(\left[I, D\right])
\end{equation*}
where %$i = 1, 2, \ldots, N_{\text{token}}$, and 
$N_{\text{token}}$ denotes the total number of prediction tokens per image. %The notation $\{f_i\}_{i=1:N_{\text{token}}}$ is a shorthand for $\{f_1,f_2,...,f_{N_{\text{token}}}\}$. 
The variable $l$ represents the network level at which Gaussians are predicted. %The backbone usually consists four levels, each processing a progressively downsampled set of tokens. 
If the second level is chosen, the feature map usually has a spatial resolution of $H/8 \times W/8$, resulting in $N_{\text{token}} = HW/64$ tokens per image. %Here, $f_{\text{enc}_i}$ represents the encoder features of an $i$-th token from the  backbone, while $f_{\text{dec}_i}$ denotes the corresponding decoder features at the same resolution. %extracted after ODIN's multi-scale deformable attention module.
% the camera's intrinsic and extrinsic parameters (from tracking system). Specifically, we concatenate the RGB-D data with the Plücker coordinates channel-wise~\cite{ziwen2024llrm, gslrm2024}, resulting in a 10-channel feature map that is fed into the model.
% The 2D backbone consists of four stages, each processing a progressively downsampled set of tokens. For Gaussian prediction, we select one stage's output and concatenate its encoded features, $f_{en_i}$, with the corresponding decoded features, $f_{de_i}$, from the multi-scale deformable attention module.
% We select one stage as the output for GS prediction and concatenate its encoded features $f_{en_i}$ from 2D backbone with the corresponding decoded features $f_{de_i}$ from the multi-scale deformable attention module. 

\noindent \textbf{Processing Prediction Tokens with Transformer.} Given the selected prediction level, we concatenate the encoder features $f_{\text{enc}_i}$ and decoder features $f_{\text{dec}_i}$ for each token $i$, and process the resulting tokens using local-attention transformer layers in the image space
%\begin{equation}
%        \{f_i\}_{i=1:N_{\text{token}}}  = \text{Transformer}(\{\left[f_{\text{enc}_i}, f_{\text{dec}_i}\right]\}_{i=1:N_{\text{token}}}), 
%\end{equation}
to obtain the final ${f_i}$ features for  the $i$-th token, integrating information from both the encoder and decoder.

\noindent \textbf{Decoding Prediction Tokens to Gaussians.}
Each output token's features, $f_i$, from the transformer layers are decoded into Gaussian parameters using Multi-Layer Perceptron (MLP):
\begin{equation*}
        \{\Delta x_i, \Delta y_i, \Delta d_i, \Delta c_i, s_i, r_i, o_i\}  = \text{MLP}(f_i),
\end{equation*}
Here, $\Delta x_i$ and $\Delta y_i$ represent the offsets from the 2D position ($x_i$, $y_i$) of the token $f_i$ in the image space, while $\Delta d_i$ is the offset for the noisy depth $d_i$ obtained from the depth image. These offsets are added to the original values, which are then backprojected into 3D space using the intrinsic and extrinsic parameters of the camera, yielding the 3D center position $\mu_i$.
Similarly, $\Delta c_i$ represents the offset for the RGB values, obtained from the downsampled image, where each token corresponds to a single pixel. Adding the offset to this value yields the final RGB color for each Gaussian.
% which are added to the original RGB value $c_i$ from the downsampled image, , yielding the final RGB value for each Gaussian. 
Besides Gaussian parameters, the mask decoder head in the segmentation model predicts token-level semantic segmentation label vector  $v_i^{\text{class}}$ for the input image, which we then assign to the associated Gaussian. Finally, each Gaussian is assigned the final feature vector $f_i$ of its corresponding token for the subsequent Gaussian refinement stage.
% \begin{equation}
%         \mu_i  = \text{3DProj}(xi + \Delta x_i, yi + \Delta y_i, di + \Delta d_i, camera),
% \end{equation}
% \begin{equation}
%         c_i  = c_{p_i} + \Delta c_i
% \end{equation}
% Where $G_i$ represents the parameters of 3D Gaussian. 
% Each token is associated with a 2D position and a depth value extracted from the depth image. 
% We use MLP layers to predict offsets for both the 2D position and the depth. These offsets are added to the original values, and the resulting 2.5D coordinates are projected into 3D space using the camera's intrinsic and extrinsic parameters, yielding the 3D center position $\mu_i \in \mathbb{R}^3$ for each Gaussian.
% Given that each token's RGB value is directly retrieved from the RGB image, we employ MLP layers to predict residuals. By adding these residuals to the original pixel RGB values, we obtain the refined RGB value $c_i \in \mathbb{R}^3$ for each Gaussian. 
% To simplify the prediction of the covariance matrix, we first predict the 3D scale vector $\mathbf{S} \in \mathbb{R}^{3}$ and 4D quaternion rotation $\mathbf{Q} \in \mathbb{R}^{4}$, which are then transformed into the 3D covariance matrix using Eq.~\ref{eq:covariance}. 

To supervise the prediction of semantic segmentation, we follow the setup in Mask2Former~\citep{cheng2021mask2former}. We denote the corresponding segmentation loss by $\mathcal{L}_{seg}$.
% Following Mask2Former~\cite{cheng2021mask2former}, we use Hungarian matching for query assignment and loss computation. Although trained only for instance segmentation, our model also outputs semantic segmentation at test time. 
In addition, we render images at $M$ supervision views—comprising the current input view and randomly selected novel views that overlap with the current input—using the predicted Gaussians from the current input, and minimize RGB-D and semantic rendering loss. For novel view supervision, we focus solely on areas visible in the input view, ensuring that the optimization process focuses on regions consistently observed across both input and novel views. 
% We use rendering and segmentation loss $\mathcal{L}_{seg}$ of mask head to train the Gaussian prediction model. 
We explain the loss functions used during training below.

\noindent \textbf{RGB Rendering Loss.}
Following previous work~\cite{gslrm2024, ziwen2024llrm}, we use a combination of the Mean Squared Error (MSE) loss and Perceptual loss: 
$
    \mathcal{L}_{rgb} = \frac{1}{M} \sum_{v=1}^M\left(\mathrm{MSE}\left(I^{gt}_{v},I^{pre}_{v}\right)
    +\lambda \cdot \mathrm{PER}\left(I^{gt}_{v},I^{pre}_{v}\right) \right),
$
where $\lambda$ is the weight for the perceptual loss.

\noindent \textbf{Depth Rendering Loss.}
For depth images, we use L1 loss:
$
    \mathcal{L}_{d} = \frac{1}{M} \sum_{v=1}^M\mathrm{L1}\left(D^{gt}_{v},D^{pre}_{v}\right).
    \label{eq:depth_loss}
$

\noindent \textbf{Semantic Rendering Loss.}
For semantic rendering, we use the cross entropy loss:
$
    \mathcal{L}_{Sem} = \frac{1}{M} \sum_{v=1}^M\mathrm{Cross\_Entropy}\left(\text{Sem}^{gt}_{v},\text{Sem}^{pre}_{v}\right).
    \label{eq:sem_loss}
$
where the rendered semantic image has N channels, each corresponding to a different semantic category.

\noindent \textbf{Overall Training Loss.} Our total loss comprises multiple rendering losses and the 
% semantic 
segmentation loss $\mathcal{L}_{seg}$:
$
    \mathcal{L} = \lambda_{rgb} \cdot \mathcal{L}_{rgb} + \lambda_{d} \cdot \mathcal{L}_{d} + \lambda_{Sem} \cdot \mathcal{L}_{Sem} + \mathcal{L}_{seg} 
    \label{eq:all_loss}
$, where we use $ \lambda_{rgb} = 1.0$, $\lambda_{d} = 1.0$ and $\lambda_{Sem} = 0.1$.
% To prevent the 3D Gaussians from becoming excessively elongated, we introduce an isotropic regularization term:
% \begin{equation}
%     L_\text{reg} = \frac{\sum_{k \in K}|s_k - \overline{s}_k|_1}{|K|},
% \end{equation}
% where $s_k\in\mathbb{R}^3$ is the scale of a 3D Gaussian, $\overline{s}_k$ is the mean sub-map scale, and $|K|$ is the number of Gaussians predicted from the input RGB-D image.

\subsubsection{Gaussian Refinement Network}
\label{sec:gs_refinement}
The previous subsection predicts Gaussian parameters from a single frame. 
% In our SLAM system, a major novelty is that we merge the predicted Gaussians from a new frame into the existing 3D map, which is maintained throughout the SLAM process.
In our SLAM system, as new frames arrive, we insert Gaussians from the frame into unmapped regions of the current 3D reconstruction. We perform co-visibility, which involves rendering a silhouette image for the new frame to identify the regions where new Gaussians should be inserted~\cite{splatam}.
To ensure that the combined set of Gaussians accurately represents the scene, we propose a novel Gaussian Refinement Network to refine both the existing Gaussians in the 3D map and the newly added ones, enabling their effective merging. The input to the network includes the features $f_i$ and 3D positions $\mu_i \in \mathbb{R}^3$ of all Gaussians from the 3D map that are visible in the new frame, as well as Gaussians from the new frame. We process these using several local-attention transformer layers with 3D neighborhoods in the world coordinate system to fuse and update the features for each Gaussian. Subsequently, MLP layers predict updates $ \Delta c_i \in \mathbb{R}^{3}$, $\Delta s_i \in \mathbb{R}^{3} $, $\Delta r_i \in \mathbb{R}^{4}$ and  $ \Delta o_i \in \mathbb{R}$ for each Gaussian. These updates refine the Gaussians to accurately render both current and previous views. %Additionally, we replace each Gaussian's features with the fused Gaussian features from the transformer layer. 
To supervise the network, we render the current view along with previous overlapping views. 
% We use a multi-stage training strategy: starting with two consecutive frames, then four, and finally eight. 
The total training loss is:% defined as follows:
\begin{equation}
\mathcal{L}_{refine} = \lambda_{rgb} \cdot \mathcal{L}_{rgb} + \lambda_{d} \cdot \mathcal{L}_{d} + \lambda_{Sem} \cdot \mathcal{L}_{Sem}
\label{eq:merge_loss}
\end{equation}
where we use $ \lambda_{rgb} = 1.0$, $\lambda_{d} = 1.0$ and $ \lambda_{Sem} = 0.1$. After Gaussian refinement, we prune Gaussians whose updated opacity falls below 0.005, effectively removing those that have become unimportant after merging. These merging-pruning steps lead to a significantly reduced number of Gaussians in the final 3D map with little performance impact.% significantly reduces the number of Gaussians in the updated 3D.

During testing time, we introduce a threshold $U$ to manage the uncertainty of each Gaussian. Once a Gaussian has been updated $U$ times by the refinement network, we consider its uncertainty sufficiently reduced and exclude them %. Consequently, in subsequent refinement steps, these Gaussians are excluded 
from further updates. We set $U=8$ in our experiments. 
\vspace{-0.05in}
\subsection{The SLAM System}
\vspace{-0.05in}
An overview of the system is summarized in Fig.~\ref{fig_pipline}. The system always maintains a set of 3D Gaussians representing the entire scene. For each new RGB-D image, the Gaussian prediction network predicts 3D Gaussian parameters, which can be rendered into high-fidelity color, depth, and semantic images. The Gaussian refinement network refines both the existing Gaussians in the 3D map and the newly added ones to accurately render both current and previous views. During testing, we occasionally run few-iteration test-time optimization and refine 3D Gaussians in the map to reflect camera pose updates from loop closure and bundle adjustment in the tracking module, while jointly optimizing the camera poses of these frames.

\vspace{-0.05in}
\subsubsection{Tracking and Global Bundle Adjustment}

\vspace{-0.05in}
\label{sec:tracking}
% For camera tacking in our SLAM system, we adopt a tracker used in GO-SLAM~\citep{zhang2023goslam} which is an enhanced version of DROID-SLAM’s tracking module~\citep{teed2021droid}. It first predicts motion in every frame. In local camera tracking, a keyframe is initialized when sufficient motion is detected, and loop closure (LC) is performed. Meanwhile, global localization performs full bundle adjustment (BA) for real-time global refinement once the system contains more than 25 keyframes. Both LC and BA help address the problem of accumulated errors and drift that can occur during the localization process.

Our main contribution is on the mapping side, hence for camera tracking in our SLAM system, we can adopt any off-the-shelf algorithm~\citep{zhang2023goslam,murai2025mast3r}.
%, which is an enhanced version of DROID-SLAM’s tracking module~\citep{teed2021droid}.
A SLAM tracking system usually consists of two components: local camera tracking and global localization. In local camera tracking, a keyframe is initialized when sufficient motion is detected, and loop closure (LC) is performed. Meanwhile, global localization performs full bundle adjustment (BA) for real-time global refinement once the system contains more than 25 keyframes. Both LC and BA help address the problem of accumulated errors and drift that can occur during the localization process.

\vspace{-0.05in}
\subsubsection{Few-Iteration Joint Gaussian$–$Pose Optimization}
\label{sec:gs_oneiter_opt}
\vspace{-0.05in}
% In our SLAM system, we aim to maintain a consistent 3D structure by refining both newly added and previously inserted Gaussians.  Our Gaussian prediction model generates optimal Gaussians for each new frame. The Gaussian Refinement refines both the existing Gaussians in the 3D map and the newly added ones to accurately render both current and previous views. 

Loop closure and bundle adjustment are essential components in SLAM systems, employed to correct accumulated drift and adjust the camera poses of previous frames.
However, these adjustments can cause Gaussians inserted based on earlier, uncorrected poses to misalign with the scene, leading to inaccurate rendering and mapping. It is crucial to implement a mechanism that updates the Gaussians in the 3D map following pose corrections. To address this issue, we propose using rendering-based optimization to update the Gaussian parameters $c_i \in \mathbb{R}^{3}$, $\mu_i \in \mathbb{R}^3$, $\mathbf{S} \in \mathbb{R}^{3} $, $\mathbf{Q} \in \mathbb{R}^{4}$ and $ o_i \in \mathbb{R}$ with only a few iterations. 
% We render RGB-D images for the top-k frames, selected based on significant pose changes. At the same time, we also use the rendering loss to optimize the camera poses of these top-k frames.
Specifically, we re-render RGB-D images for the top-$k$ frames, selected based on significant pose changes, and jointly minimize a rendering loss with respect to both the Gaussian parameters and the camera poses of these frames.
The optimized camera poses are then fed back into the tracking system, and used for loop closure and bundle adjustment.
This joint optimization keeps the 3D map consistent with the corrected poses while also leveraging the full 3D map to further improve pose accuracy.
To enhance the efficiency of this optimization, we employ the batch rendering technique from~\cite{ye2024gsplatopensourcelibrarygaussian}. We omit semantic image rendering to improve system efficiency. For few-iteration optimization, we add a SSIM term to the RGB loss, following~\cite{kerbl3Dgaussians}:
\begin{equation}
\begin{aligned}
    \mathcal{L}_{opt}  & =  \frac{1}{M} \sum_{i'=1}^M\big(\lambda_{rgb}\cdot \big((1 - \lambda) \cdot \mathrm{L1}\left(I^{gt}_{i'},I^{pre}_{i'}\right) 
    +  \\
    & \lambda \big(1 - \mathrm{SSIM}(I^{gt}_{i'},I^{pre}_{i'}) \big) + \lambda_{d}\cdot\mathrm{L1}(D^{gt}_{i'},D^{pre}_{i'}) \big)
\label{eq:optimiza_loss}
\end{aligned}
\end{equation}
where $\lambda$ is set to $0.2$ for all experiments. We perform 1–5 iterations in our experiments.

%% file: sec/4_experiment.tex
\section{Experiments}
\subsection{Experimental Setup}
\noindent \textbf{Training Settings.} We train our Gaussian prediction and refinement networks \textbf{entirely} on RGB-D videos from the real ScanNet datasets. We exclude the six standard SLAM test scenarios and use all remaining training and validation scenes, supervising with 20 common semantic classes. 
For ScanNet++, we exclude the SLAM test scenarios and use all other available training scenes, supervising with the 100 most common semantic classes. 
We adopt AutoFocusFormer~\citep{autofocusformer} as the backbone for both Mask2Former and Gaussian prediction, using the second stage of the backbone as the prediction stage. Following the low-to-high resolution curriculum of~\citep{ziwen2024llrm}, we train the Gaussian prediction network in three stages with input resolutions of 256×256, 480×480, and 640×480. In the first two stages, images are resized such that the shorter side is 256 or 480 pixels and then center-cropped to a square. 
On ScanNet++, we use resolutions of 256x384. 
For the refinement network, which processes multiple consecutive frames, we adopt a progressive training schedule: beginning with two frames, then four, and finally eight.

\noindent \textbf{Evaluation Datasets and Settings.} During testing, we evaluate our method on six real-world scenes from ScanNet~\citep{dai2017scannet}, which are commonly used as SLAM benchmarks, and six real-world scenes from ScanNet++~\citep{yeshwanth2023scannet++}. ScanNet++ validation set contains 11 scenes with continuous camera trajectories, but SplaTAM and SGS-SLAM \textbf{completely fail} on 5 of them. Hence we only use the 6 scenes they can finish running, so that their numbers are not infinitely bad. Note that our approach can successfully run on \textbf{all} 11 scenes. 
Additionally, we perform zero-shot experiments on real scenes from NYUv2~\citep{Silberman:ECCV12} and TUM RGB-D~\citep{sturm12iros}.
% , as well as synthetic scenes from Replica~\citep{straub2019replica}.
For ScanNet, NYUv2, and TUM RGB-D, we evaluate rendering performance on every 5th frame of each sequence. For ScanNet++, we evaluate all training views of each RGB-D video and additionally hold out the novel views provided by the dataset.

\noindent \textbf{Evaluation Metrics.} We use PSNR, Depth-L1~\citep{Zhu2022CVPR}, SSIM~\citep{1284395}, and LPIPS~\citep{zhang2018perceptual} to evaluate the reconstruction and rendering quality. We additionally report reconstruction metrics such as Accuracy, Completion, Completion Ratio ($<$7cm) and F1 ($<$7cm) in the appendix.
For GS-based SLAM methods, we also report the number of Gaussians. For semantic segmentation, we report the mean Intersection over Union (mIoU). To evaluate the accuracy of the camera pose, we adopt the average absolute trajectory error (ATE RMSE)~\citep{6385773}.

\noindent \textbf{Baselines.} We compare our method against several state-of-the-art approaches: NeRF-based SLAM methods, including NICE-SLAM~\citep{Zhu2022CVPR}, GO-SLAM~\citep{zhang2023goslam}, and Point-SLAM~\citep{Sandström2023ICCV}; 
% the semantic NeRF-based SLAM method DNS-SLAM~\citep{li2023dnsslamdenseneural}; 
3D Gaussian-based SLAM methods such as SplaTAM~\citep{keetha2024splatam}, RTG-SLAM~\citep{peng2024rtgslam}, and GS-ICP SLAM~\citep{ha2024rgbdgsicpslam}; and semantic 3D Gaussian-based SLAM methods, including SGS-SLAM~\citep{li2024sgs}, GS$^3$LAM~\citep{li2024gs3lam} and OVO-Gaussian-SLAM~\citep{martins2024ovo}.
%For the GO-SLAM~\citep{zhang2023goslam} comparison, we ran a low-resolution (240 $\times$ 320) version on ScanNet to match their original setup. 
%For other baselines, we used the original ScanNet resolution of 480 $\times$ 640.
% We exclude SNI-SLAM~\citep{zhu2024sni} from our comparisons because their evaluation utilizes the training set of the segmentation model on the Replica dataset, and they have not released code for the ScanNet dataset.
% Regarding SGS SLAM~\citep{li2024sgs}, their approach employs a test-time optimization method with ground truth semantic labels for supervision. 
% DNS-SLAM, 
SGS-SLAM, GS$^3$LAM and OVO-Gaussian-SLAM are the only semantic SLAM methods available for comparison since the code is not available for other semantic SLAM approaches. Note that SGS-SLAM~\citep{li2024sgs} and GS$^3$LAM~\citep{li2024gs3lam} employ test-time optimization using ground truth semantic labels on the test set. SGS-SLAM and GS$^3$LAM have been shown to outperform all other existing semantic SLAM methods~\citep{zhu2024sni, li2023dnsslamdenseneural}.
To ensure a fair comparison and simulating SLAM applications in real-world scenarios where ground truth semantic labels are unavailable, we trained a 2D segmentation model using a Swin backbone~\citep{liu2021Swin} with Mask2Former~\citep{cheng2021mask2former} on ScanNet and ScanNet++, following the same training strategy as our model, and used predicted semantic labels to supervise SGS-SLAM nad GS$^3$LAM.
% their SLAM system.

\subsection{Results}

\begin{table}[t]
\caption{ \textbf{Rendering Performance on ScanNet.} Values are averaged across the test scenes. Best results are highlighted as \colorbox{colorFst}{\bf first}, \colorbox{colorSnd}{second}. %All methods are run at $640 \times 480$ and $320 \times 240$ resolution.
GS Num represents the number of 3D Gaussians included in the scene after mapping is complete.} 
\vskip -0.1in
\label{tab:scannet_rendering}
\begin{center}
\begin{small}
\resizebox{\linewidth}{!}{%
\begin{tabular}{lllccccc}
\toprule
\bf Res &\bf Method & \bf PSNR$\uparrow$ & \bf SSIM$\uparrow$ & \bf LPIPS$\downarrow$ & \bf Depth L1$\downarrow$ & \bf GS Num$\downarrow$ \\
\midrule
 & NICE-SLAM      & 17.54 & 0.621 & 0.548 & -      & -      \\
 & Point-SLAM     &  19.82 &  0.751 & 0.514 & -      & -      \\
 & SplaTAM        & 18.99 & 0.702 & 0.364 & 7.21   & 2466k  \\
$640 \times 480$ & RTG SLAM       & 12.75 & 0.372 & 0.761 & 97.56  & 1229k  \\
 & GS-ICP SLAM    & 14.73 & 0.645 & 0.684 & 103.31  & 2565k  \\
 & SGS SLAM    & 15.89 & 0.594 & 0.615 & 11.83  & 2114k  \\
 & GS$^3$LAM    & 20.67 & 0.796 & \fs 0.288 & 11.88  & 2154k  \\
 & \bf GS4 (Ours, 1 iter)   & \nd 22.62 & \nd 0.851 &  0.338 & \nd 6.46 &  \nd 321k  \\
 & \bf GS4 (Ours, 5 iters)  & \fs 24.55 & \fs 0.885 & \nd 0.299 & \fs 4.86 & \fs 224k \\
% \cmidrule(lr){1-8}
\midrule
 & GO-SLAM       & 18.21 & 0.657 & 0.553 & 18.14 & -      \\
 $320 \times 240$& \bf GS4 (Ours, 1 iter) & \nd 22.54 & \nd 0.885 & \nd 0.240 & \nd 5.98 & \nd 162k \\
 & \bf GS4 (Ours, 5 iters)  & \fs 24.31 & \fs 0.921 &  \fs 0.196 & \fs 4.93 & \fs 124k \\
 \bottomrule
\end{tabular}
}
\end{small}
\end{center}
\vskip -0.2in
\end{table}

\noindent \textbf{Rendering and Reconstruction Performance.}
In Table~\ref{tab:scannet_rendering}, we evaluate the rendering and reconstruction performance of our method on ScanNet. This is a difficult task compared to the synthetic data where neural RGB-D SLAM methods usually show strong results, because inevitably inaccurate ground truth camera poses and depths make optimization much harder than completely clean synthetic datasets. Compared to existing dense neural RGB-D SLAM methods, our approach achieves state-of-the-art performance on PSNR, SSIM, and Depth L1 metrics. Specifically, our method surpasses the runner-up, GS$^3$LAM \citep{li2023dnsslamdenseneural}, by 3.88 dB in PSNR (a 18.8$\%$ percent improvement).
and 0.089 in SSIM (a 11.2$\%$ percent improvement).
Furthermore, %when compared to other GS-based SLAM methods such as SplaTAM, GS-ICP SLAM, and SGS-SLAM, 
our approach utilizes approximately \textbf{10x} fewer Gaussians than the baselines. This efficiency highlights the effectiveness of our method in achieving high-quality scene representation with reduced computational complexity. 

In Table~\ref{tab:scannet_rendering}, we ran our method with $240 \times 320$ input to compare against GO-SLAM which shares the same tracking method as ours but renders at the same low resolution. GS4 maintains the same PSNR and depth prediction quality as its high resolution version and significantly outperforms GO-SLAM across all metrics.

\begin{figure*}[htb]
\centering
{\footnotesize
\setlength{\tabcolsep}{1pt}
\renewcommand{\arraystretch}{0.9}
\newcommand{\sz}{0.15}
\begin{tabular}{ccccccc}
\rotatebox[origin=c]{90}{GS-ICP SLAM} & 
\raisebox{-0.5\height}{\includegraphics[width=\sz\linewidth]{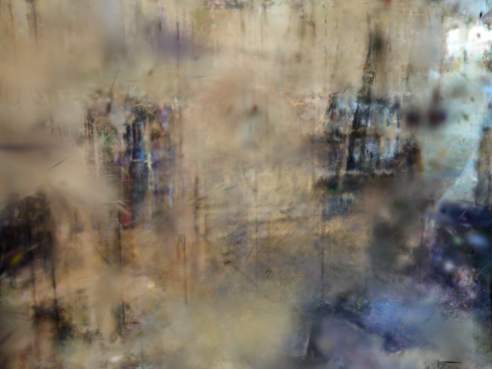}} & 
\raisebox{-0.5\height}{\includegraphics[width=\sz\linewidth]{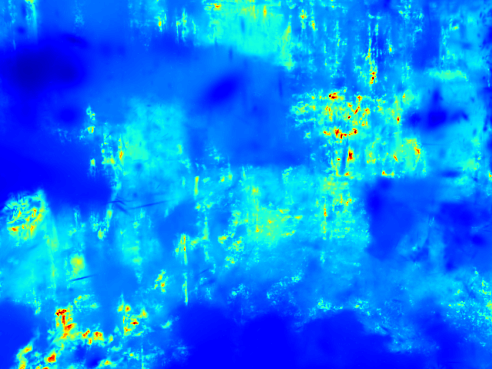}} 
&
\raisebox{-0.5\height}{\includegraphics[width=\sz\linewidth]{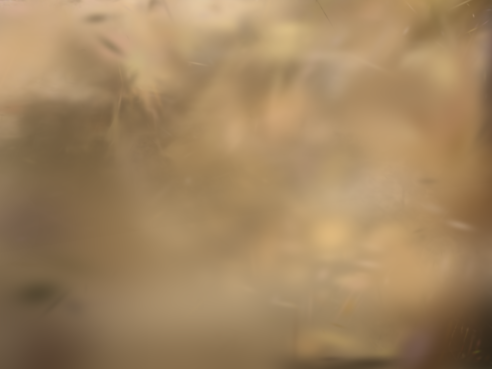}} &
\raisebox{-0.5\height}{\includegraphics[width=\sz\linewidth]{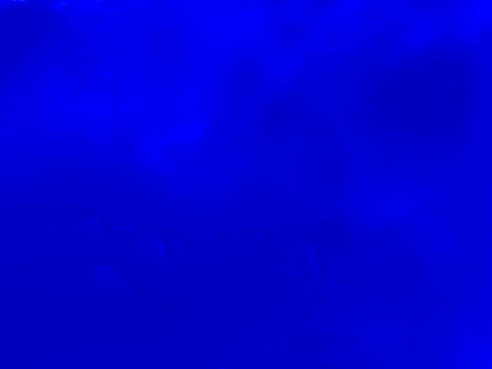}}
& 
\raisebox{-0.5\height}{\includegraphics[width=\sz\linewidth]{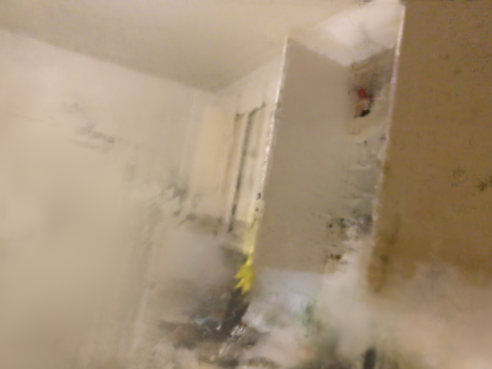}} & 
\raisebox{-0.5\height}{\includegraphics[width=\sz\linewidth]{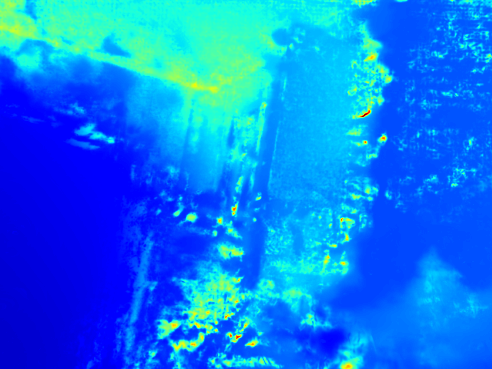}} 
\\
\rotatebox[origin=c]{90}{SplaTAM} & 
\raisebox{-0.5\height}{\includegraphics[width=\sz\linewidth]{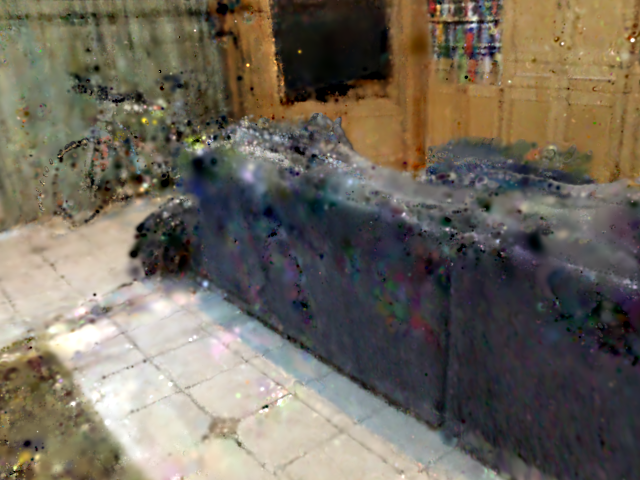}} & 
\raisebox{-0.5\height}{\includegraphics[width=\sz\linewidth]{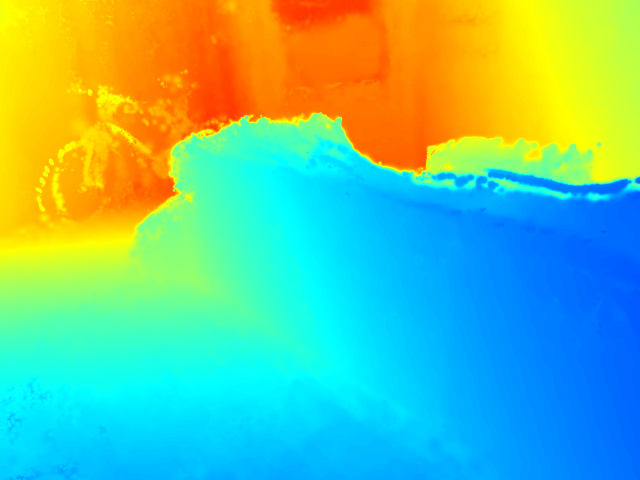}} 
&
\raisebox{-0.5\height}{\includegraphics[width=\sz\linewidth]{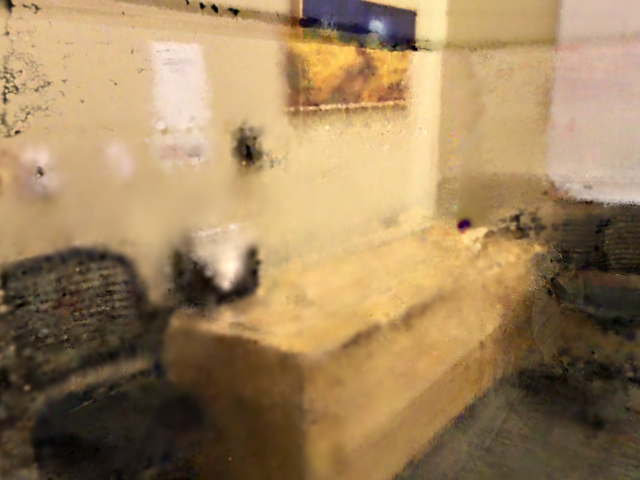}} &
\raisebox{-0.5\height}{\includegraphics[width=\sz\linewidth]{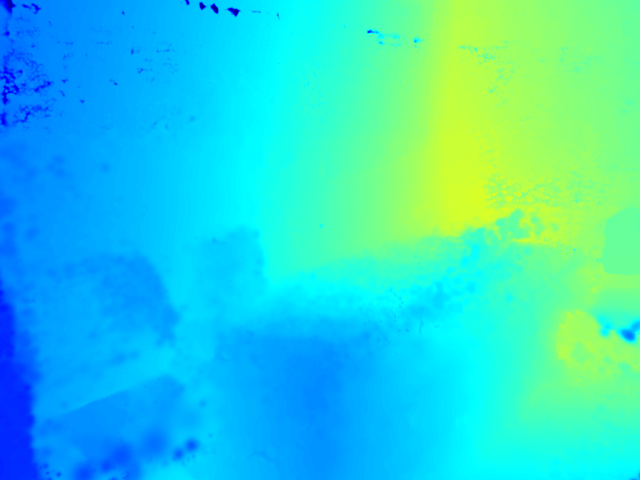}}
& 
\raisebox{-0.5\height}{\includegraphics[width=\sz\linewidth]{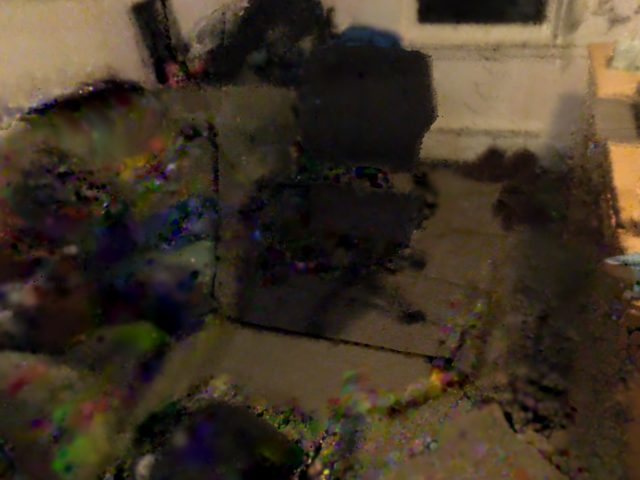}} & 
\raisebox{-0.5\height}{\includegraphics[width=\sz\linewidth]{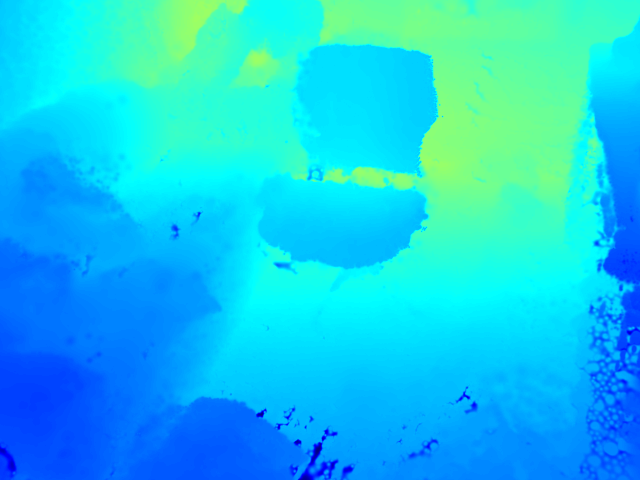}} 
\\
\rotatebox[origin=c]{90}{SGS SLAM} & 
\raisebox{-0.5\height}{\includegraphics[width=\sz\linewidth]{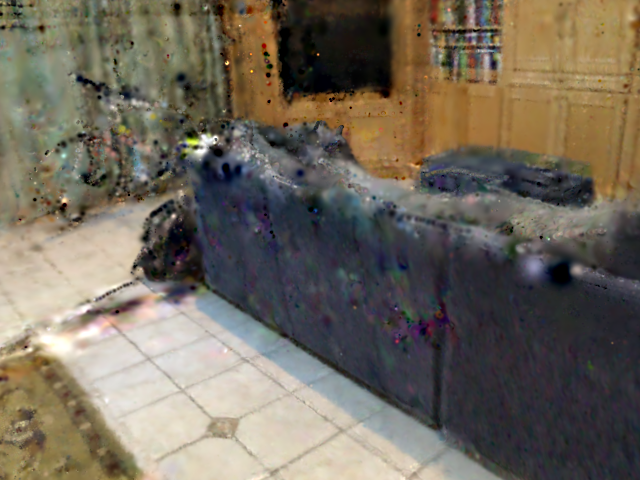}} & 
\raisebox{-0.5\height}{\includegraphics[width=\sz\linewidth]{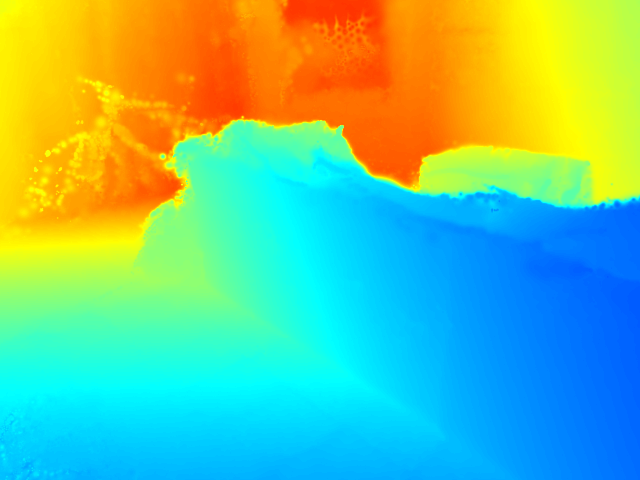}} 
&
\raisebox{-0.5\height}{\includegraphics[width=\sz\linewidth]{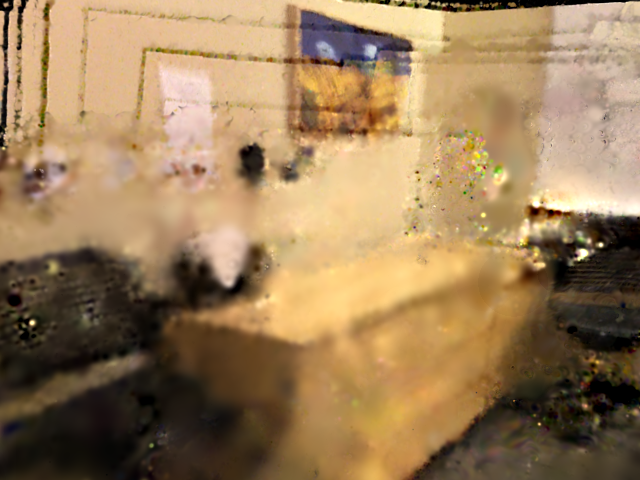}} &
\raisebox{-0.5\height}{\includegraphics[width=\sz\linewidth]{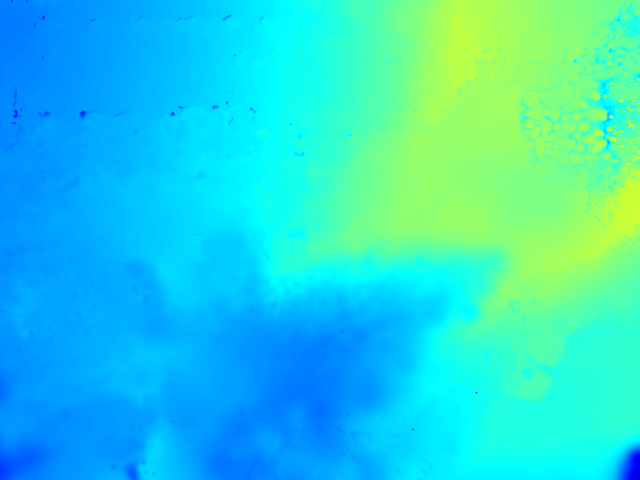}}
& 
\raisebox{-0.5\height}{\includegraphics[width=\sz\linewidth]{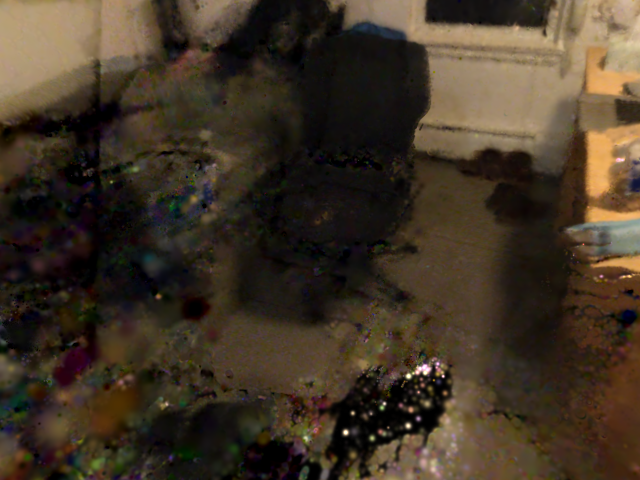}} & 
\raisebox{-0.5\height}{\includegraphics[width=\sz\linewidth]{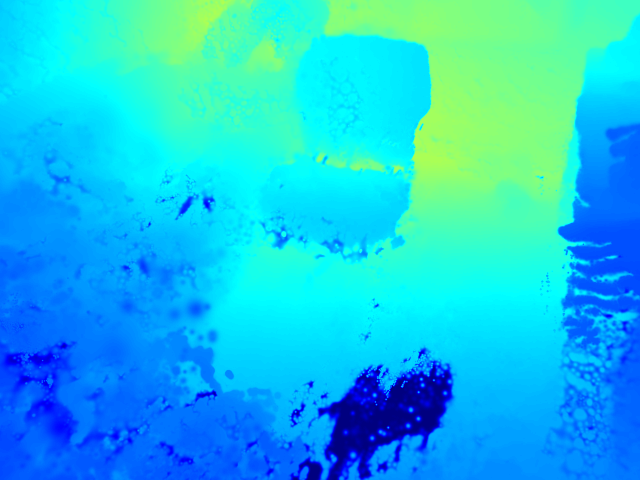}} 
\\
\rotatebox[origin=c]{90}{GS$^3$LAM} & 
\raisebox{-0.5\height}{\includegraphics[width=\sz\linewidth]{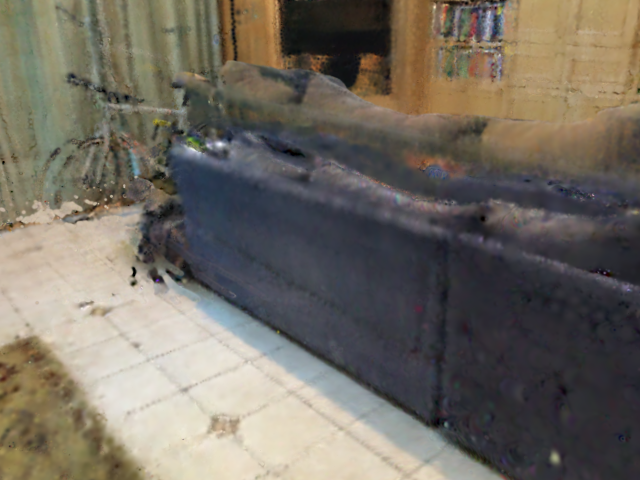}} & 
\raisebox{-0.5\height}{\includegraphics[width=\sz\linewidth]{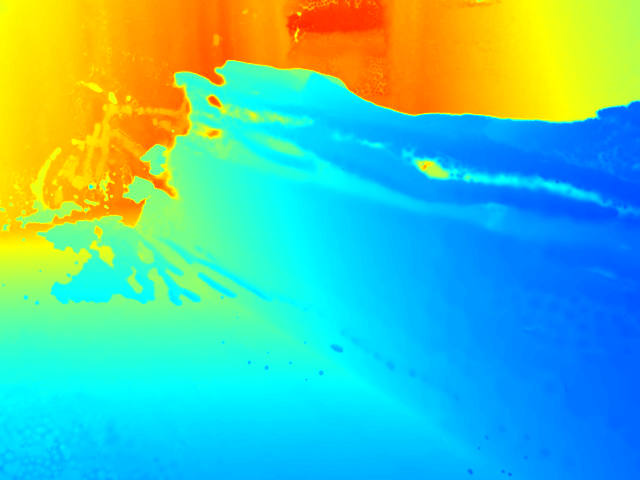}} 
&
\raisebox{-0.5\height}{\includegraphics[width=\sz\linewidth]{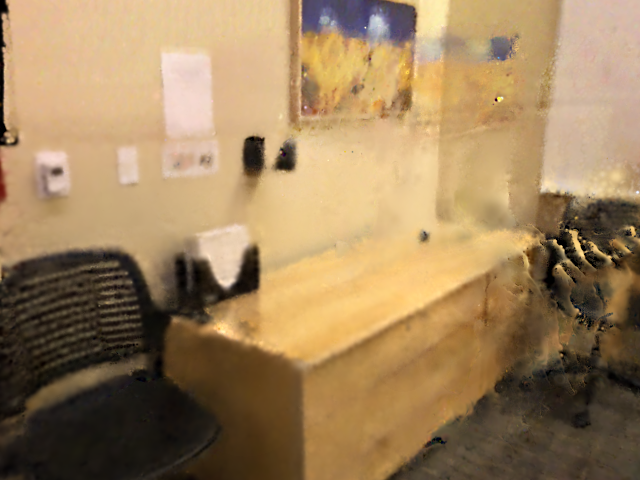}} &
\raisebox{-0.5\height}{\includegraphics[width=\sz\linewidth]{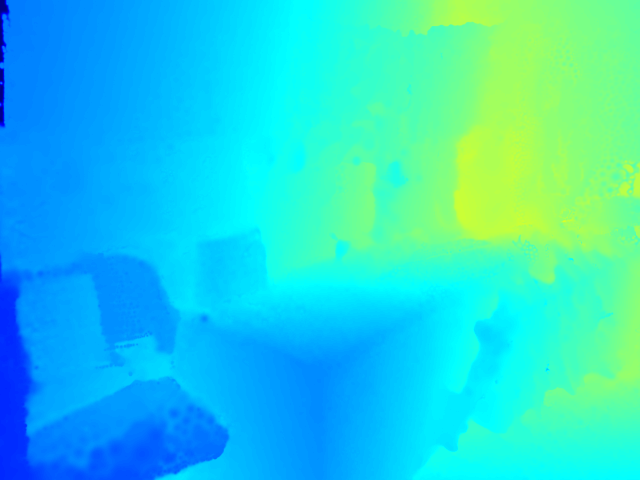}}
& 
\raisebox{-0.5\height}{\includegraphics[width=\sz\linewidth]{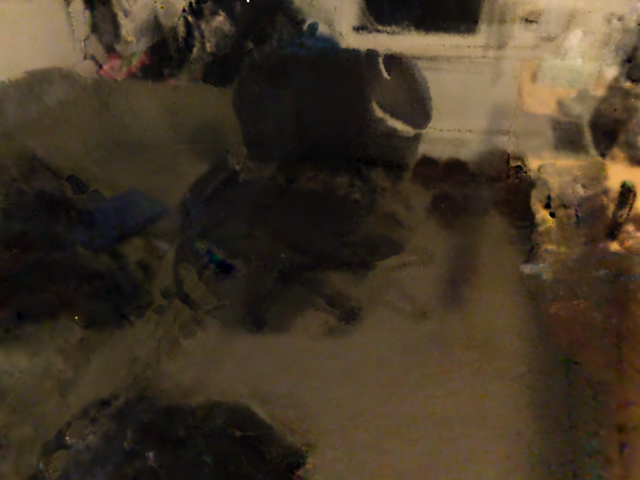}} & 
\raisebox{-0.5\height}{\includegraphics[width=\sz\linewidth]{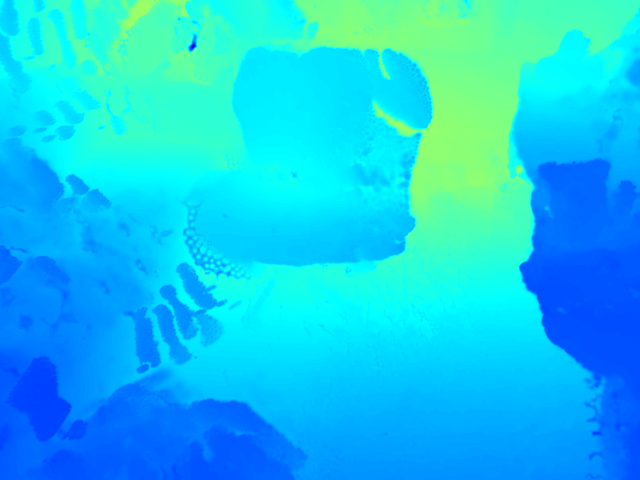}} 
\\
\rotatebox[origin=c]{90}{\textbf{GS4 (Ours)}} & 
\raisebox{-0.5\height}{\includegraphics[width=\sz\linewidth]{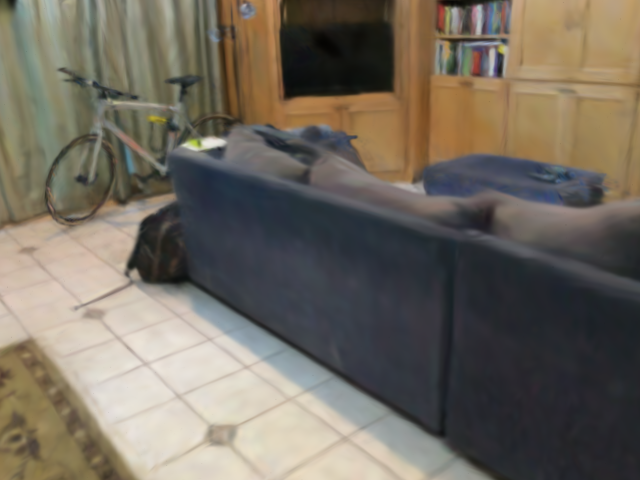}} & 
\raisebox{-0.5\height}{\includegraphics[width=\sz\linewidth]{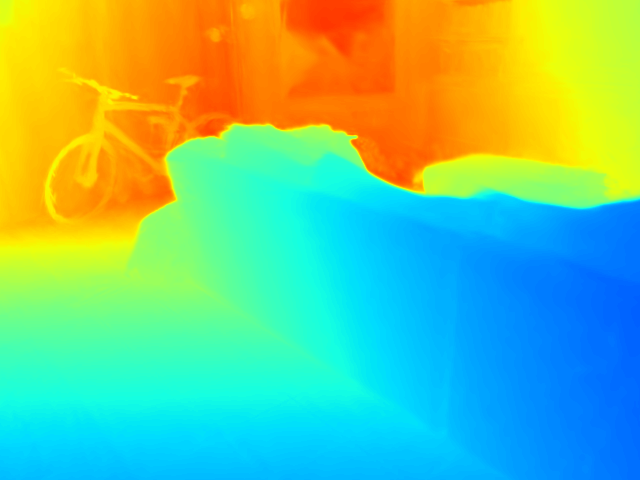}} 
&
\raisebox{-0.5\height}{\includegraphics[width=\sz\linewidth]{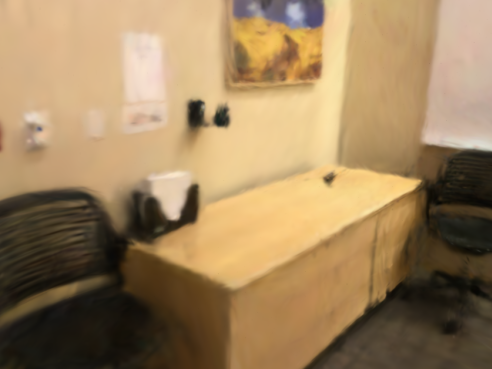}} &
\raisebox{-0.5\height}{\includegraphics[width=\sz\linewidth]{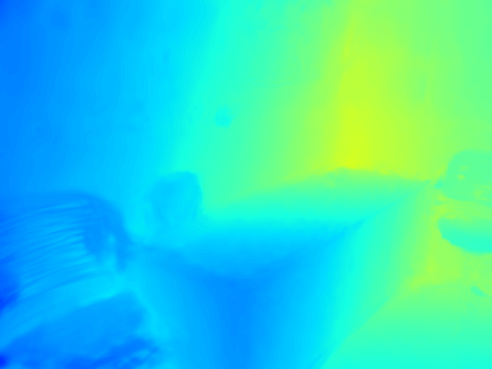}}
& 
\raisebox{-0.5\height}{\includegraphics[width=\sz\linewidth]{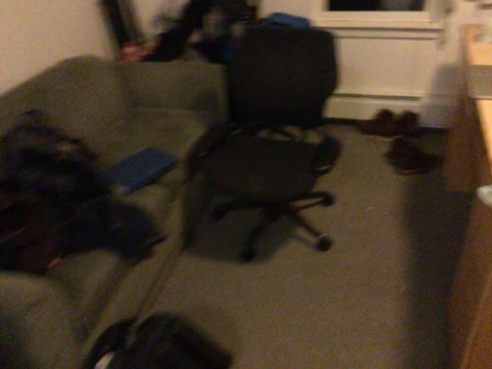}} & 
\raisebox{-0.5\height}{\includegraphics[width=\sz\linewidth]{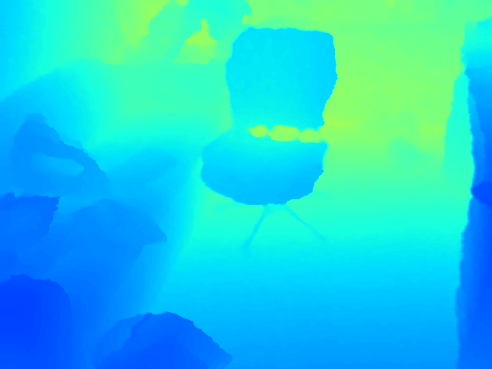}} 
\\
\rotatebox[origin=c]{90}{Ground Truth} & 
\raisebox{-0.5\height}{\includegraphics[width=\sz\linewidth]{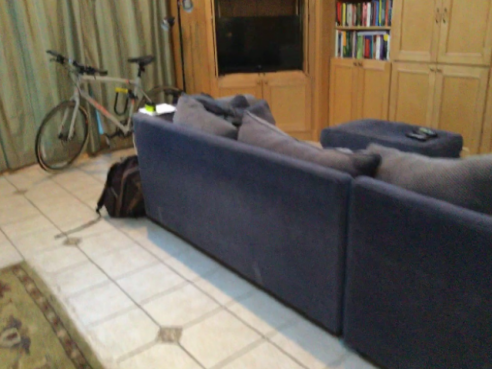}} & 
\raisebox{-0.5\height}{\includegraphics[width=\sz\linewidth]{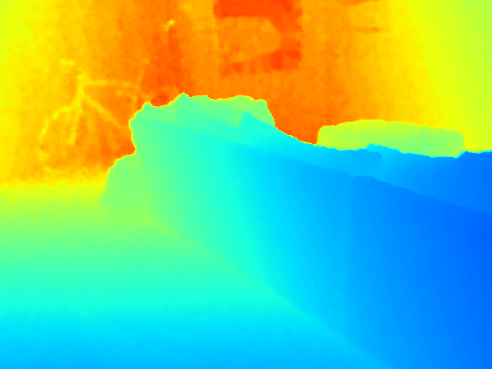}} 
&
\raisebox{-0.5\height}{\includegraphics[width=\sz\linewidth]{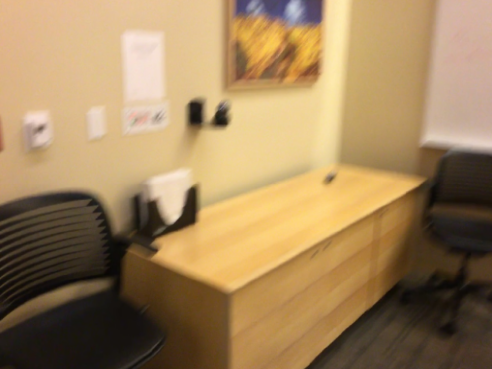}} &
\raisebox{-0.5\height}{\includegraphics[width=\sz\linewidth]{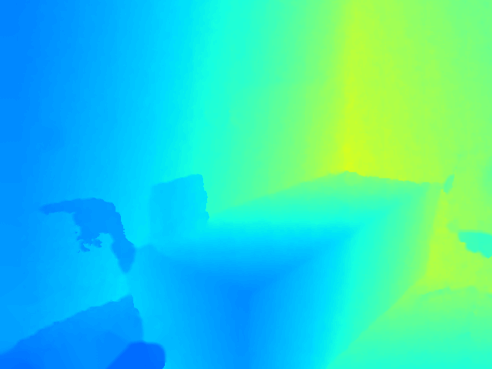}}
& 
\raisebox{-0.5\height}{\includegraphics[width=\sz\linewidth]{figures/scannet_vis/0207/1700.png}} & 
\raisebox{-0.5\height}{\includegraphics[width=\sz\linewidth]{figures/scannet_vis/0207/depth_1700.png}} 
\\
& \multicolumn{2}{c}{\texttt{scene0000}} & \multicolumn{2}{c}{\texttt{scene0169}} & \multicolumn{2}{c}{\texttt{scene0207}} 
\\

\end{tabular}
}
\vskip -0.1in
\caption{\textbf{Renderings on ScanNet}. Our method, \textbf{GS4}, renders color \& depth for views with fidelity significantly better than all approaches.%comparable to the ground truth. 
%It can also be observed that other GS-based SLAM method, especially for GS-ICP SLAM and SGS SLAM, fail to provide good renderings on all images. The failure can be attributed to the failure of tracking as shown in \cref{tab:scannet_tracking}.
}
\label{fig:scannet_vis}
\vskip -0.1in
\end{figure*}

Fig.~\ref{fig:scannet_vis} shows visual results of RGB and depth rendering. Our method demonstrates superior performance than other GS-SLAM methods. %scene rendering. 
Notably, sometimes the depth maps of our approach even turn out to be better than the noisy ground truth depth inputs. For instance, in the first two columns, our method delivers a more contiguous and complete rendering of the bicycle tires. Similarly, in the middle two columns, we reconstruct the chair's backrest nearly entirely, whereas the GT depth data lacks this detail.

\noindent \textbf{Semantic Performance.} 
% Since SGS-SLAM~\citep{li2024sgs} employs ground truth labels for supervision, for a fair comparison and simulation of SLAM application in real scenarios (No GT labels), we trained 2D segmentation model and use the predicted labels for supervision. 
In Table~\ref{tab:scannet_semantic}, we present both 2D rendering and 3D mean Intersection over Union (mIoU) scores across the six ScanNet test scenes. For 3D mIoU evaluation, we first align the reconstructed map with the ground-truth mesh and then use 3D neighborhood voting to assign predicted labels. 
% Our method outperforms the previous runner-up, GS$^3$LAM, by 19.19$\%$ in 3D mIoU. In terms of 2D mIoU, our method surpasses the state-of-the-art semantic 3Dgs-based SLAM approach, GS$^3$LAM, by 17.01$\%$. Qualitative comparisons are included in the appendix. 
Our method outperforms the previous runner-up, GS$^3$LAM, by 20.42$\%$ in 3D mIoU and by 7.29$\%$ in 2D mIoU. Qualitative comparisons are provided in the appendix.

\begin{table}[t]
% \vskip -0.1in
\caption{\textbf{Semantic Performance across ScanNet Test Scenes
}
}
\vskip -0.15in
\label{tab:scannet_semantic}
\begin{center}
\resizebox{0.99\linewidth}{!}{%
  \begin{small}
\begin{tabular}{lccccc}
\toprule
\bf Methods & SGS SLAM & OVO-Gaussian-SLAM & GS$^3$LAM & \bf GS4 (Ours, 1 iter) & \bf GS4 (Ours, 5 iters) \\
\midrule

 mIoU(2D) & 37.20 & \redx & 56.42 & \fs 63.71 & \nd 62.10 \\ 

 mIoU(3D) & 18.87 & 32.58 & 34.42 & \fs 54.84 & \nd 53.61 \\ 
\bottomrule
\end{tabular}
\end{small}
  }
\end{center} 
\vskip -0.2in
\end{table}

\noindent \textbf{Tracking Performance.} 
Table~\ref{tab:scannet_tracking} shows the tracking results. Our method uses the same tracking algorithm as GO-SLAM, which is significantly better than other GS-based SLAM methods. %reduces the trajectory error over the prior state-of-the-art dense baseline, Point-SLAM, by more than 35$\%$, decreasing from 10.70 cm to 6.98 cm. Other GS-based SLAM methods, such as GS-ICP SLAM and SGS SLAM, exhibit significantly poorer tracking performance.

\begin{table}[t]
\begin{center}
\caption{\textbf{Tracking Performance on ScanNet Test Scenes}. The average values are reported. GS4 uses the same tracking algorithm as GO-SLAM hence the numbers are almost the same.}
\label{tab:scannet_tracking}
% \resizebox{1.0\linewidth}{!}{%
\resizebox{\linewidth}{!}{
\begin{small}
\begin{tabular}{lcccccc}
\toprule
\bf Metric & NICE-SLAM & Point-SLAM & SplaTAM & RTG SLAM & GS-ICP SLAM \\
\midrule
ATE RMSE [cm]$\downarrow$ & \nd 10.70 & 12.19 & 11.88 & 144.52 & NaN \\
\midrule
\bf Metric & SGS SLAM & GS$^3$LAM & GO-SLAM & \bf GS4 (Ours, 1 iter) & \bf GS4 (Ours, 5 iters) \\
ATE RMSE [cm]$\downarrow$ & 40.97 & 30.88 & \fs 7.00 & \fs 6.97 & \fs 6.95 \\
\bottomrule
\end{tabular}
\end{small}
}
\end{center}
\vskip -0.2in
\end{table}

\begin{table}[t]
% \vskip -0.2in
\caption{\textbf{Average Runtime on ScanNet Test Scenes}}
\vskip -0.2in
\label{tab:scannet_fps}
% \vskip -0.2in
\begin{center}
\resizebox{0.99\linewidth}{!}{%
  \begin{small}
\begin{tabular}{lcccc}
\toprule
\bf Methods    & Point-SLAM & SplaTAM & RTG-SLAM & GS-ICP SLAM \\
\midrule
FPS $\uparrow$ & 0.05 & 0.23 & 1.01 & \fs 3.62 \\
\midrule
\bf Methods    & SGS-SLAM & GS$^3$LAM & GS4 (ours, 1 iter) & GS4 (ours, 5 iters) \\
FPS $\uparrow$ & 0.17  & 0.12 & \nd 2.85 & 1.82 \\
\bottomrule
\end{tabular}
\end{small}
  }
\end{center}
\vskip -0.3in
\end{table}

\noindent \textbf{Runtime Comparison.}
Table~\ref{tab:scannet_fps} presents a runtime comparison of our method against the baselines at the $640 \times 480$ resolution, conducted on an Nvidia RTX TITAN. FPS is calculated by dividing the total number of frames by the total time to represent the overall system performance. While GS-ICP SLAM is faster than ours, its rendering and tracking performance is significantly worse (Table \ref{tab:scannet_rendering} and  \ref{tab:scannet_tracking}). Our approach is \textbf{12x} faster than SplaTAM, \textbf{17x} faster than  SGS-SLAM, and \textbf{24x} faster than GS$^3$LAM. Notably, even with this exceptional speed, our approach maintains superior map quality and outperforms other methods.

\begin{table}[t]
\centering
\caption{\textbf{Comparison of averaged performance on ScanNet++ at 256$\times$384 resolution.} The 2D rendering metrics are averaged over both training and novel views. GS Num denotes the number of 3D Gaussians after mapping is complete, and FPS denotes the runtime in frames per second.}
\label{tab:train_novel_views}
\resizebox{0.99\linewidth}{!}{%
\begin{tabular}{lcccccccc}
\toprule
 \bf Method & \multicolumn{4}{c}{\bf Rendering Metrics} & \bf mIoU (3D)$\uparrow$ & \bf GS Num$\downarrow$ & \bf ATE RMSE$\downarrow$ & \bf FPS$\uparrow$ \\
\cmidrule(lr){2-5}
& PSNR$\uparrow$ & SSIM$\uparrow$ & LPIPS$\downarrow$ & mIoU (2D)$\uparrow$ 
  & & \\
\midrule
SplaTAM & \nd 20.34 & \fs 0.758 & \nd 0.331 & \redx &  \redx & \nd 1899k & 1588.95 & \nd 0.17 \\
SGS-SLAM & 15.59 & 0.533 & 0.470 & \nd 15.35 & \nd 0.58 & 2374k & \nd 1315.89 & 0.13 \\
\bf GS4 (Ours) & \fs 21.06 & \nd 0.740 & \fs 0.281 & \fs 18.92 & \fs 13.80 & \fs 85k & \fs 5.37 & \fs 1.60\\
\bottomrule
\end{tabular}
}
\end{table}

\noindent \textbf{ScanNet++ Experiments.} Table~\ref{tab:train_novel_views} reports 2D rendering performance averaged over both training and novel views, as well as 3D semantic, tracking, and runtime comparisons against the baselines at a resolution of 256$\times$384, averaged over six scenes. Our method achieves state-of-the-art performance on both 2D and 3D tasks, outperforming the runner-up by 0.72dB in PSNR (3.5$\%$ relative improvement), 0.05 in LPIPS (15.1$\%$ relative improvement), 3.57$\%$ in 2D mIoU, and 13.22$\%$ in 3D mIoU, while using only \textbf{3.0$\%$–4.5$\%$} of the Gaussians required by other methods.
Compared to SGS-SLAM, although it achieves the second-best 2D mIoU, its 3D mIoU drops to just 0.58$\%$, indicating poor generalization to the full 3D scene. In addition to delivering superior performance across both 2D and 3D metrics, our method also exhibits substantially higher efficiency, running \textbf{9×} faster than SplaTAM and \textbf{12×} faster than SGS-SLAM.

\begin{table}[t]
% \vskip -0.05in
\caption{ \textbf{Zero-shot Rendering Performance on NYUv2 and TUM-RGBD.} Values are averaged across the test scenes. GS Num represents the number of 3D Gaussians in the scene after mapping.}
\vskip -0.15in
\label{tab:zeroshot_rendering}
\begin{center}
\resizebox{0.99\linewidth}{!}{%
\begin{small}
\resizebox{\linewidth}{!}{%
\begin{tabular}{lllccccc}
\toprule
\bf Dataset & \bf Res &\bf Method & \bf PSNR$\uparrow$ & \bf SSIM$\uparrow$ & \bf LPIPS$\downarrow$ & \bf GS Num$\downarrow$ \\
\midrule
& & SplaTAM  & 18.86 & 0.692 & 0.372   & 1236k \\
NYUv2 & $640 \times 480$ & RTG-SLAM  & 11.84 & 0.221 & 0.703 & \nd 807k \\
& & SGS-SLAM  & \nd 19.32 & \nd 0.708 & \nd 0.357 &   1108k \\
& & \bf GS4 (Ours) & \fs 22.24 & \fs 0.866 & \fs 0.254 & \fs 298k   \\
\midrule
& & SplaTAM & \fs 22.76 & \nd 0.891 & \fs 0.182 &   803k \\
TUM RGBD & $640 \times 480$ & RTG-SLAM & 19.75 & 0.769 & 0.395 & \nd 198k \\
& & SGS-SLAM  & 22.44 & 0.876 & \nd 0.184 &   735k \\
 &  & \bf GS4 (Ours) & \nd 22.70 & \fs 0.903 & 0.191 & \fs 166k \\
 \bottomrule
\end{tabular}
}
\end{small}
}
\end{center}
\vskip -0.2in
\end{table}

\noindent \textbf{Zero-shot Experiments.} In Table~\ref{tab:zeroshot_rendering}, we report quantitative zero-shot results. For NYUv2, the numbers are averaged over three scenes, and for TUM-RGBD, they are also averaged over three scenes. Per-scene results are provided in the appendix. 
On NYUv2, our method outperforms all other GS-based SLAM approaches across all rendering metrics, achieving a 15$\%$–20$\%$ relative improvement over the runner-up while using significantly fewer Gaussians. Qualitative comparisons are also provided in the appendix. 
On TUM-RGBD, our method outperforms the baselines in terms of SSIM and the number of Gaussians, and closely matches the best performance in other metrics, despite relying primarily on a feed-forward model trained on ScanNet.

\noindent \textbf{Ablation Study.} We conduct an ablation study using all ScanNet test scenes, as shown in Table~\ref{tab:scannet_ablation}. The results demonstrate that both the Gaussian Refinement Network and the Few-Iteration Joint Gaussian–Pose Optimization are critical to the performance of GS4. Additionally, Gaussian pruning significantly reduces the number of Gaussians without sacrificing accuracy. 

\begin{table}[t]
% \vskip -0.05in
\begin{center}
\caption{\textbf{Ablation on ScanNet (averaged over test scenes)}}
\label{tab:scannet_ablation}
\resizebox{0.99\linewidth}{!}{%
  \begin{small}
\begin{tabular}{@{}cccccccc@{}}
\toprule
\bf Design Choice  & PSNR [dB]$\uparrow$ & SSIM$\uparrow$ & LPIPS$\downarrow$ & Depth L1$\downarrow$ & mIoU$\uparrow$ & ATE $\downarrow$ & Gs Num$\downarrow$\\
\midrule
GS Prediction  & 15.05 & 0.461 & 0.662 & 33.17 & 40.11 & 6.99 & 133k \\
+ GS Refinement  & 16.1 & 0.556 & 0.584 & 29.81 & 44.65 & 7.00 & 581k \\
+ 1-Iter. Optimization & 22.72 & 0.851 & 0.337 & 6.24 & 63.58 & 6.97 & 666k \\
+ GS Pruning (Full SLAM) & 22.62 & 0.851 & 0.338 & 6.46 & 63.71 & 6.97 & 321k \\
% + Pose opt & 22.62 & 0.851 & 0.338 & 6.46 & 63.71 & 6.97 & 321k \\
 \bottomrule
\end{tabular}
\end{small}
}
\end{center}
\vskip -0.2in
\end{table}

\noindent \textbf{Ablation Study of Few-Iteration Joint Gaussian–Pose Optimization.} In Sec.~\ref{sec:gs_oneiter_opt}, we propose to perform few-iteration joint Gaussian–pose optimization and feed the refined poses back into the tracking system to further improve pose accuracy. In Table~\ref{tab:scannet_ablation_pose_opt}, we ablate (i) Gaussian-only optimization and (ii) joint Gaussian–pose optimization with pose feedback into the tracking system. For the rendering metrics, joint Gaussian–pose optimization yields slight but consistent improvements over Gaussian-only optimization on both datasets. For tracking accuracy, we observe that pose feedback is particularly effective when high-quality ground truth is available, as in ScanNet++, whose RGB-D images are of much higher quality than those in ScanNet.

\begin{table}[t]
% \vskip -0.05in
\begin{center}
\caption{\textbf{Ablation of joint Gaussian–pose optimization on ScanNet and ScanNet++ (averaged over test scenes).} Results are reported using 5 iterations of the few-iteration joint Gaussian–pose optimization.}
\vskip -0.05in
\label{tab:scannet_ablation_pose_opt}
\resizebox{0.99\linewidth}{!}{%
  \begin{small}
\begin{tabular}{@{}cccccccc@{}}
\toprule
\bf Dataset & \bf Optimization & \bf PSNR$\uparrow$ & \bf SSIM$\uparrow$ & \bf LPIPS$\downarrow$ & \bf GS Num$\downarrow$  & \bf ATE RMSE$\downarrow$ \\
\midrule
ScanNet & Gaussian & 24.39 & 0.881 & 0.302 & 225k & 6.94 \\
& Gaussian–Pose & 24.55 & 0.885 & 0.299 & 224k & 6.95 \\
\midrule
ScanNet++ & Gaussian  & 20.88 & 0.736 & 0.282 & 90k & 6.34 \\
& Gaussian–Pose & 21.06 & 0.740 & 0.281 & 85k & 5.37 \\
 \bottomrule
\end{tabular}
\end{small}
}
\end{center}
\vskip -0.2in
\end{table}

%% file: sec/5_conclusion.tex
\section{Conclusion}

We present GS4, a novel SLAM system that incrementally constructs and updates a 3D semantic scene representation from a RGB-D video with a learned generalizable network. Our novel Gaussian refinement network and few-iteration joint Gaussian-pose optimization significantly improve the performance of our approach. Our experiments demonstrate state-of-the-art semantic SLAM performance on the ScanNet benchmark while running 10x faster and using 10x less Gaussians than baselines. The model also showed strong generalization capabilities through zero-shot transfer to the NYUv2 and TUM RGB-D datasets. In future work, we will further improve the computational speed of GS4 and explore options for a pure RGB-based SLAM approach.

%% file: sec/X_suppl.tex
% % CVPR 2026 Paper Template; see https://github.com/cvpr-org/author-kit

% \documentclass[10pt,twocolumn,letterpaper]{article}

% %%%%%%%%% PAPER TYPE  - PLEASE UPDATE FOR FINAL VERSION
% % \usepackage{cvpr}              % To produce the CAMERA-READY version
% \usepackage[review]{cvpr}      % To produce the REVIEW version
% % \usepackage[pagenumbers]{cvpr} % To force page numbers, e.g. for an arXiv version

% % Import additional packages in the preamble file, before hyperref
% \input{preamble}

% % It is strongly recommended to use hyperref, especially for the review version.
% % hyperref with option pagebackref eases the reviewers' job.
% % Please disable hyperref *only* if you encounter grave issues, 
% % e.g. with the file validation for the camera-ready version.
% %
% % If you comment hyperref and then uncomment it, you should delete *.aux before re-running LaTeX.
% % (Or just hit 'q' on the first LaTeX run, let it finish, and you should be clear).
% \definecolor{cvprblue}{rgb}{0.21,0.49,0.74}
% \usepackage[pagebackref,breaklinks,colorlinks,allcolors=cvprblue]{hyperref}

% %%%%%%%%% PAPER ID  - PLEASE UPDATE
% \def\paperID{8958} % *** Enter the Paper ID here
% \def\confName{CVPR}
% \def\confYear{2026}

% \begin{document}

% % \maketitle
\clearpage
\appendix
% \setcounter{page}{1}
% \maketitlesupplementary
% \section*{\center \huge Appendix}
\noindent                                                                \textbf{\huge Appendix}
\vspace{0.25in}

\section{More Experimental Setup}
For ScanNet, we evaluate on six scenes—scene0000, scene0059, scene0106, scene0169, scene0181, and scene0207—which are commonly used by existing SLAM methods. For ScanNet++, we use six scenes in total: 8b5caf3398, 3f15a9266d, e7af285f7d, 99fa5c25e1, 09c1414f1b, and 9071e139d9. For NYUv2, we use the sequences bedroom$\_$0051$\_$out, dining$\_$room$\_$0031$\_$out, and student$\_$lounge$\_$0001$\_$out. For TUM RGB-D, we evaluate on rgbd$\_$dataset$\_$freiburg1$\_$desk, rgbd$\_$dataset$\_$freiburg2$\_$xyz, and rgbd$\_$dataset$\_$freiburg3$\_$long$\_$office$\_$household.

\section{Ablation on semantic head} 
We present the semantic head ablation in Tab.~\ref{tab:scannet_ablation_semantic}. The versions with and without semantic prediction exhibit similar rendering performance, suggesting that semantics do not noticeably change reconstruction quality. This trend is consistent with the behavior of SGS-SLAM and SplaTAM. SGS-SLAM is built on top of SplaTAM by adding semantic colors to Gaussians and introducing a semantic loss for tracking and mapping. However, on the ScanNet, ScanNet++ and TUM RGB-D datasets, incorporating semantics actually degrades performance, with SGS-SLAM performing worse than SplaTAM. Taken together, these results indicate that incorporating semantic information does not inherently affect SLAM tracking or reconstruction quality.
Nevertheless, we argue that having a unified backbone that jointly supports both SLAM and semantics remains valuable, as it allows a single model to deliver both accurate geometry and rich semantic understanding of the scene.

\begin{table}[htb]
% \vskip -0.05in
\begin{center}
\caption{\textbf{Ablation of semantic prediction head on ScanNet (averaged over test scenes).} Results are reported using 5 iterations of the few-iteration joint Gaussian–pose optimization.}
\vskip -0.05in
\label{tab:scannet_ablation_semantic}
\resizebox{0.99\linewidth}{!}{%
  \begin{small}
\begin{tabular}{@{}cccccc@{}}
\toprule
\bf Semantic Head & \bf PSNR$\uparrow$ & \bf SSIM$\uparrow$ & \bf LPIPS$\downarrow$ & \bf GS Num$\downarrow$  & \bf ATE RMSE$\downarrow$ \\
\midrule
False & \nd 24.53 & \fs 0.885 & \fs 0.292 & \nd254k & \nd 6.96 \\
True & \fs 24.55 & \fs 0.885 & \nd 0.299 & \fs 224k & \fs 6.95 \\
 \bottomrule
\end{tabular}
\end{small}
}
\end{center}
\vskip -0.2in
\end{table}

\section{More Results on GS4 with Different Prediction Stages}
In the main paper, we use the second stage of the backbone as the default prediction stage. Here, we additionally report results using the third and fourth stages of the backbone as the prediction stage in Table~\ref{tab:scannet_rendering_postopt} and Table~\ref{tab:nyu_tum_rendering_postopt}. Using the fourth stage substantially reduces the number of Gaussians in the scene—by roughly \textbf{$ 10 \times$ fewer}—while only slightly degrading the rendering quality and the semantic performance. As shown in Table~\ref{tab:scannet_rendering_postopt}, the fourth stage achieves PSNR that is comparable to the second stage, despite using only one tenth of the Gaussians, highlighting a favorable trade-off between reconstruction efficiency and fidelity.
We also observe that, under the same prediction stage, using a lower input resolution leads to only a small drop in rendering performance, but quite significantly degraded semantic performance. This is because our model is ultimately trained at a higher resolution of $640 \times 480$, and downscaling the input reduces the amount of fine-grained semantic detail available per frame. Consequently, the per-frame semantic predictions become less accurate, which accumulates over the sequence and results in a noticeable drop in semantic performance at the scene level.

\begin{table}[b]
\caption{ \textbf{Rendering Performance on ScanNet.} Values are averaged across the test scenes. Results are reported using 5 iterations of the few-iteration joint Gaussian–pose optimization. GS Num represents the number of 3D Gaussians included in the scene after mapping is complete.} 
\vskip -0.1in
\label{tab:scannet_rendering_postopt}
\begin{center}
\begin{small}
\resizebox{\linewidth}{!}{%
\begin{tabular}{llccccccc}
\toprule
\bf Res & \bf Method & \bf PSNR$\uparrow$ & \bf SSIM$\uparrow$ & \bf LPIPS$\downarrow$ & \bf GS Num$\downarrow$ & \bf mIOU(2D)$\uparrow$ & \bf ATE $\downarrow$\\
\midrule
  & Splat-SLAM  & \redx & \redx &  \redx & \redx   & \redx & \redx\\
$640 \times 480$ & \bf GS4 (Ours, 2nd stg)  & \nd 24.55 & \fs 0.885 & \fs 0.299  &  224k & \fs 62.10 & \nd 6.95 \\
   & \bf GS4 (Ours, 3rd stg)   & \fs 24.56 & \nd 0.875 & \nd 0.344 & \nd 69k & \nd 60.34 & \fs 6.94 \\
   & \bf GS4 (Ours, 4th stg)   &   24.21 & 0.859 &  0.397 & \fs 23k &  60.12 & 6.97  \\
   \midrule
  & Splat-SLAM  & 20.67 & 0.684 & 0.438  &  87k & \redx & 7.60 \\
 $320 \times 240$ & \bf GS4 (Ours, 2nd stg)  & 22.43 & \fs 0.921 & \fs 0.196  & 124k & \fs 52.85 & \nd 6.94 \\
   & \bf GS4 (Ours, 3rd stg) & \fs 24.24 & \nd 0.911 & \nd 0.245 &\nd  37k & 47.55 &\fs 6.93 \\
   &  \bf GS4 (Ours, 4th stg) & \nd 23.65 & 0.891 & 0.323 & \fs 12k & \nd 48.71 & 6.95 \\
 \bottomrule
\end{tabular}
}
\end{small}
\end{center}
\vskip -0.2in
\end{table}

\section{Comparison with SLAM Methods Using Offline Post-Optimization Refinement}
We omitted Splat-SLAM~\cite{sandstrom2024splat}, MonoGS~\cite{Matsuki:Murai:etal:CVPR2024} and OVO-Gaussian-SLAM~\cite{martins2024ovo} from the main paper because these methods perform an additional \textbf{26k} optimization iterations on the full map \textbf{on the complete video} after SLAM has finished. This heavy post-optimization step substantially boosts their final reconstruction quality (around \textbf{$+4$ dB PSNR}), but is against the mentality of SLAM for generating online and incremental outputs. It is not directly comparable to our setting, where we report results of online SLAM without such long global refinement (roughly \textbf{5 minutes} of extra optimization per scene). To ensure a fair comparison, we therefore disable these final refinement iterations on the full map and evaluate the baselines under a setting that matches ours more closely.

In Table~\ref{tab:scannet_rendering_postopt}, we report a comparison between our method and Splat-SLAM on ScanNet at two different resolutions, $640 \times 480$ and $320 \times 240$. The higher resolution is included because all other baselines report rendering metrics at $640 \times 480$, and this is also the resolution used in our main experiments. However, Splat-SLAM \textbf{fails} under this setting on ScanNet, making a direct comparison at full resolution impossible. To still provide a fair and informative comparison, we additionally evaluate at the lower resolution of $320 \times 240$, which is the default configuration for Splat-SLAM on ScanNet. This allows us to compare performance at the resolution where Splat-SLAM is stable, while simultaneously highlighting that it does not robustly handle the higher-resolution setting used by other methods.
At $320 \times 240$, our method consistently outperforms Splat-SLAM when using either the second or the fourth backbone stage for prediction. With the second stage, we surpass Splat-SLAM by 1.76 dB in PSNR (8.5$\%$), 0.237 in SSIM (34.6$\%$), and 0.242 in LPIPS (55.3$\%$). Even when using the fourth stage, we still outperform Splat-SLAM while using only 13.8$\%$ of its Gaussians, demonstrating a significantly better trade-off between quality and efficiency.

In Table~\ref{tab:nyu_tum_rendering_postopt}, we report a comparison between our method and MonoGS, OVO-Gaussian-SLAM, and Splat-SLAM on the NYUv2 and TUM RGB-D datasets. MonoGS and OVO-Gaussian-SLAM share exactly the same reconstruction metrics because OVO-Gaussian-SLAM is built on top of the MonoGS system and simply adds semantic labels using CLIP features to predict a label for each Gaussian; both methods share the same mapping and tracking pipeline for RGB-D reconstruction. It is clear that, without the long global refinement stage, our method surpasses MonoGS, OVO-Gaussian-SLAM, and Splat-SLAM on most metrics on both NYUv2 and TUM RGB-D.
In conclusion, our method outperforms these baselines that rely on post-optimization refinement on most metrics, while using substantially fewer Gaussians.

\begin{table}[t]
\caption{ \textbf{Rendering Performance on NYUv2 and TUM-RGBD using $640 \times 480$ resolution.} Values are averaged across the test scenes. Results are reported using 5 iterations of the few-iteration joint Gaussian–pose optimization. GS Num represents the number of 3D Gaussians included in the scene after mapping is complete.} 
\vskip -0.2in
\label{tab:nyu_tum_rendering_postopt}
\begin{center}
\begin{small}
\resizebox{\linewidth}{!}{%
\begin{tabular}{lllcccc}
\toprule
\bf Dataset & \bf Method & \bf PSNR$\uparrow$ & \bf SSIM$\uparrow$ & \bf LPIPS$\downarrow$ & \bf GS Num$\downarrow$ &  \bf ATE $\downarrow$\\
\midrule
  & MonoGS    & 12.88 & 0.505 & 0.550 & 110k & - -  \\
  & OVO-Gaussian-SLAM    & 12.88 & 0.505 & 0.550 & 110k & - -  \\
  & Splat-SLAM    & \fs 23.30 & 0.754 & \nd 0.332 & 85k &  -- \\
 NYUv2 & \bf GS4 (Ours, 2nd stg)  & \nd 22.24 & \fs 0.866 & \fs 0.254  & 298k & - - \\
  & \bf GS4 (Ours, 3rd stg)   & 21.86 & \nd 0.826 & 0.398 & \nd 72k & -- \\
   & \bf GS4 (Ours, 4th stg)   & 20.94 &  0.766 & 0.530 & \fs 18k & - - \\
   \midrule
   & MonoGS    & 17.78 & 0.718 & 0.315 & \nd 37k & 1.52  \\
  & OVO-Gaussian-SLAM    & 17.78 & 0.718 & 0.315 & \nd 37k & 1.52  \\
  & Splat-SLAM    & \fs 22.84 & 0.780 & 0.287 & 62k & \fs 1.1 \\
 TUM  & \bf GS4 (Ours, 2nd stg)  & \nd 22.70 & \fs 0.903 & \fs 0.191 & 166k & \nd 1.41 \\
  & \bf GS4 (Ours, 3rd stg)   &  22.32 & \nd 0.876 & \nd 0.273 & 41k & 1.42 \\
   & \bf GS4 (Ours, 4th stg)   & 21.23 & 0.833 & 0.392 & \fs 12k & 1.42 \\
 \bottomrule
\end{tabular}
}
\end{small}
\end{center}
\vskip -0.2in
\end{table}

\section{Do 3DGS-Based Methods Improve with a Stronger Tracking Module from GO-SLAM?}

We adopt off-the-shelf tracking algorithms~\cite{zhang2023goslam, murai2025mast3r} in our system. Specifically, we use the GO-SLAM tracking module on ScanNet, NYUv2, and TUM RGB-D, and the Mast3r-SLAM tracking module on ScanNet++. To clarify whether other 3DGS-based methods could also benefit from improved tracking when equipped with a similar module, and to enable a more fair comparison with existing 3DGS-based SLAM systems, we compare against Splat-SLAM and MonoGS. Splat-SLAM adopts the MonoGS mapping pipeline but integrates it into an enhanced DROID-SLAM tracking framework, which is  similar to our use of GO-SLAM-style tracking (which itself is derived from DROID-SLAM) on ScanNet, NYUv2, and TUM RGB-D.

We report the results of Splat-SLAM, MonoGS, and our method on these three datasets in Table~\ref{tab:scannet_rendering_postopt} and Table~\ref{tab:nyu_tum_rendering_postopt}. As expected, with a stronger tracking module, Splat-SLAM achieves noticeably better rendering performance than MonoGS, confirming that improved tracking alone can boost reconstruction quality for 3DGS-based systems. However, even under this stronger tracking setting, our method still substantially outperforms Splat-SLAM on most metrics, indicating that our backbone design and Gaussian prediction strategy provide additional gains beyond what can be achieved by improved tracking alone.

To further examine whether stronger tracking could also benefit semantic 3DGS-based SLAM, we attempted to inject GO-SLAM poses into GS$^3$LAM by using GO-SLAM’s local pose estimates to initialize GS$^3$LAM’s tracking. In practice, this integration turned out to be unstable: GS$^3$LAM’s internal tracking and mapping pipeline is tightly coupled to its own pose-update dynamics, and the externally initialized poses conflicted with its optimization trajectory. This mismatch caused the estimated trajectory to drift abruptly, after which the system kept spawning large numbers of Gaussians to compensate for the growing misalignment. As a result, memory usage rapidly escalated and the run eventually terminated with an out-of-memory failure. 
% \begin{table}[htb]
% \centering
% % \resizebox{0.99\linewidth}{!}{%
% \caption{Comparison between Swin and AFF backbones on different metrics.}
% \label{tab:swin_aff}
% \resizebox{0.7\linewidth}{!}{%
% \begin{tabular}{lcccccc}
% \toprule
% Backbone & PSNR & SSIM & LPIPS & Depth L1 & mIoU & GS Num \\
% \midrule
% GS4 (Swin, 1 iter) & 22.50 & 0.846 & 0.356 & 6.338 & 64.4 & 361k \\
% GS4 (AFF, 1 iter) & 22.61 & 0.851 & 0.335 & 6.558 & 63.0 & 355k \\
% \bottomrule
% \end{tabular}
% }
% \end{table}

\section{Reconstruction Results on ScanNet}
In Table~\ref{tab:reconstruction_scannet}, we use the metrics including Accuracy (Acc.), Completion (Comp.), Completion Ratio[$<$7cm] and F-Score[$<$7cm] to evaluate the scene geometry on ScanNet. The definitions of the evaluation metrics are detailed in
Table~\ref{tab:reconstruction_metrics}. GS4 outperforms all baselines in terms of completion and F-score.

\begin{table}[h]
\caption{Metric definitions. p and p$^*$are the reconstructed and ground truth point clouds}
\vskip -0.05in
\label{tab:reconstruction_metrics}
\centering
\resizebox{0.99\linewidth}{!}{%
  \begin{small}
\begin{tabular}{l c}
\toprule
\textbf{3D Metric} & \textbf{Formula} \\
\midrule
Acc & $\displaystyle \text{mean}_{p \in P} \left( \min_{p^* \in P^*} \| p - p^* \| \right)$ \\[8pt]
Comp & $\displaystyle \text{mean}_{p^* \in P^*} \left( \min_{p \in P} \| p - p^* \| \right)$ \\[8pt]
Completion Ratio[$<$7cm] & $\displaystyle \text{mean}_{p^* \in P^*} \left[ \min_{p \in P} \| p - p^* \| < 0.07 \right]$ \\[8pt]
F-score[$<$7cm] & $\displaystyle \frac{2 \times \text{mean}_{p \in P} \left[ \min_{p^* \in P^*} \| p - p^* \| < 0.07 \right] \times \displaystyle \text{mean}_{p^* \in P^*} \left[ \min_{p \in P} \| p - p^* \| < 0.07 \right]}{\text{mean}_{p \in P} \left[ \min_{p^* \in P^*} \| p - p^* \| < 0.07 \right] + \displaystyle \text{mean}_{p^* \in P^*} \left[ \min_{p \in P} \| p - p^* \| < 0.07 \right]}$ \\
\bottomrule
\end{tabular}
\end{small}
}
\end{table}

\begin{table}[htb]
\caption{Reconstruction metrics on ScanNet}
\vskip -0.1in
\label{tab:reconstruction_scannet}
\begin{center}
\resizebox{0.99\linewidth}{!}{%
  \begin{small}
\begin{tabular}{@{}cccccc@{}}
\toprule
\bf Methods   &  Acc. $\downarrow$  & Comp. $\downarrow$  & Comp. Ratio (\textless7cm) $\uparrow$ & F-Score (\textless7cm) $\uparrow$ & GS Num$\downarrow$ \\
\midrule
SplaTAM & \fs 8.10 & \nd 5.58 & \nd 76.34 & \nd 75.95 & 2466k \\
    % \hline
RTG-SLAM & 99.80 & 47.44 & 24.61 & 16.69 & \nd 1229k \\
SGS-SLAM & 17.11 & 13.75 & 55.01 & 55.26 & 2114k \\
    % \hline
GS$^3$LAM & 11.83 & 4.62 & 82.82 & 71.92 & 2154k \\
% OVO (Gaussian-SLAM) &  &  &  &  & &\\
GS4  & \nd 8.48 & \fs 3.87 & \fs 89.09 & \fs 77.44 & \fs 224k \\
\bottomrule
\end{tabular}
\end{small}
}
\end{center}
\vskip -0.2in
\end{table}

\section{Qualitative Comparison for Semantic Segmentation.} 
\begin{figure}[h]
\centering
{\footnotesize
\setlength{\tabcolsep}{1pt}
\renewcommand{\arraystretch}{0.9}
\newcommand{\sz}{0.31}
\begin{tabular}{cccc}
\rotatebox[origin=c]{90}{SGS SLAM} & 
\raisebox{-0.5\height}{\includegraphics[width=\sz\linewidth]{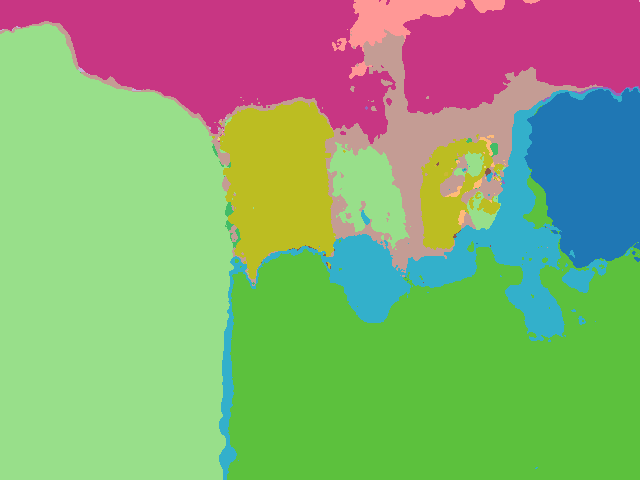}} 
& 
\raisebox{-0.5\height}{\includegraphics[width=\sz\linewidth]{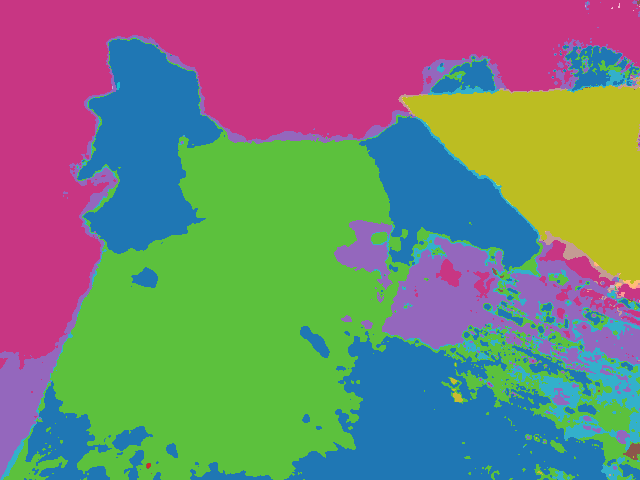}} 
&
\raisebox{-0.5\height}{\includegraphics[width=\sz\linewidth]{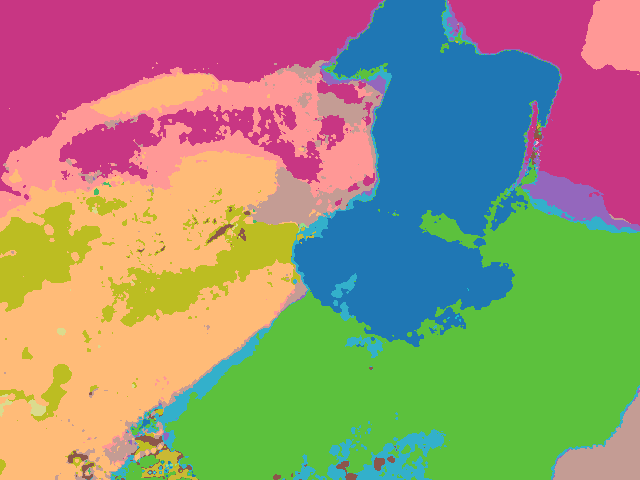}} 
\\
\rotatebox[origin=c]{90}{GS$^3$LAM} & 
\raisebox{-0.5\height}{\includegraphics[width=\sz\linewidth]{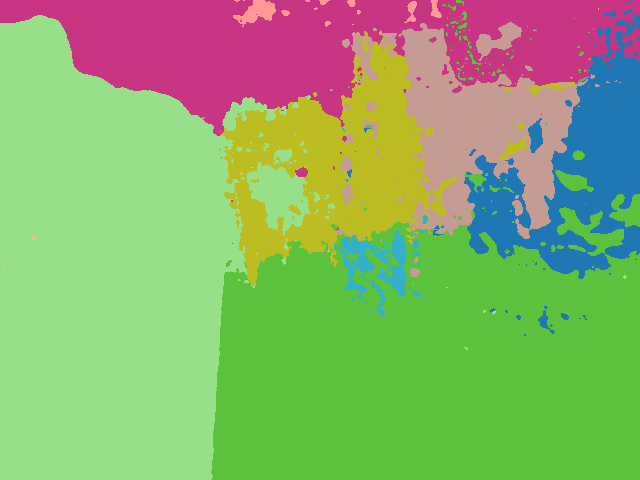}} 
& 
\raisebox{-0.5\height}{\includegraphics[width=\sz\linewidth]{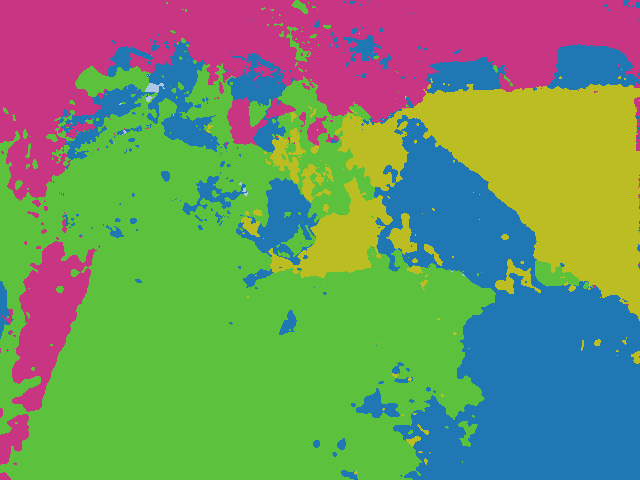}} 
&
\raisebox{-0.5\height}{\includegraphics[width=\sz\linewidth]{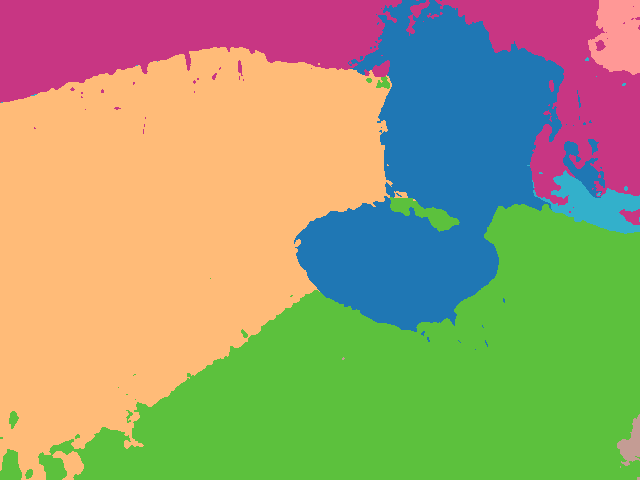}} 
\\
\rotatebox[origin=c]{90}{\textbf{GS4 (Ours)}} & 
\raisebox{-0.5\height}{\includegraphics[width=\sz\linewidth]{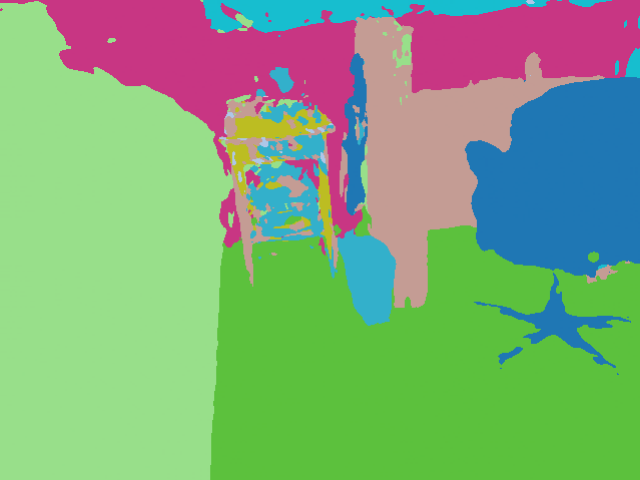}} 
& 
\raisebox{-0.5\height}{\includegraphics[width=\sz\linewidth]{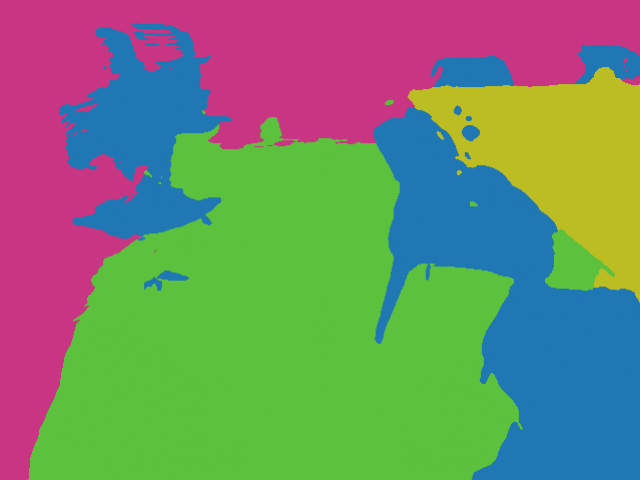}} 
&
\raisebox{-0.5\height}{\includegraphics[width=\sz\linewidth]{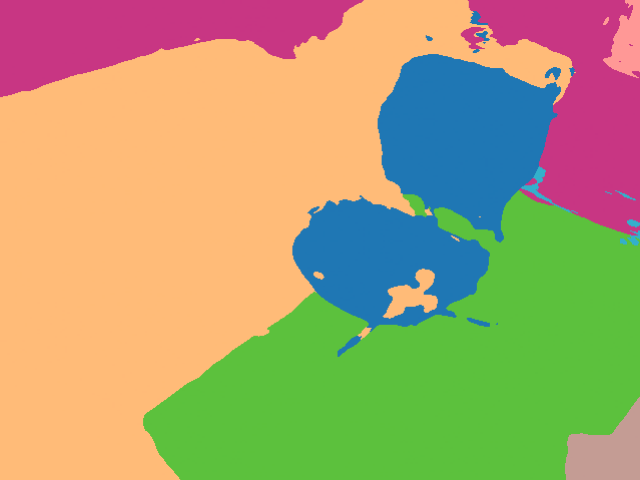}} 
\\
\rotatebox[origin=c]{90}{GT (labels)} & 
\raisebox{-0.5\height}{\includegraphics[width=\sz\linewidth]{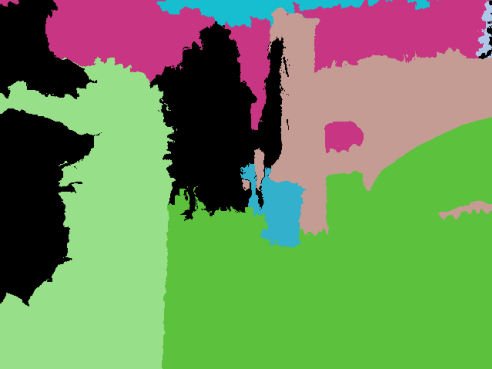}} 
& 
\raisebox{-0.5\height}{\includegraphics[width=\sz\linewidth]{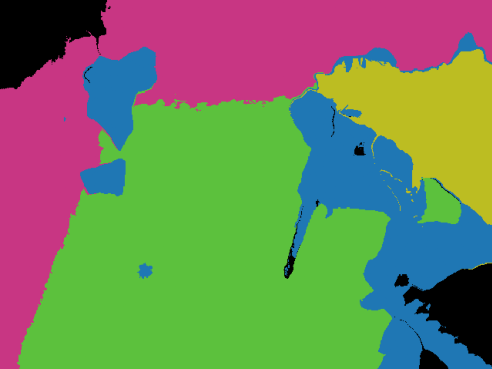}} 
&
\raisebox{-0.5\height}{\includegraphics[width=\sz\linewidth]{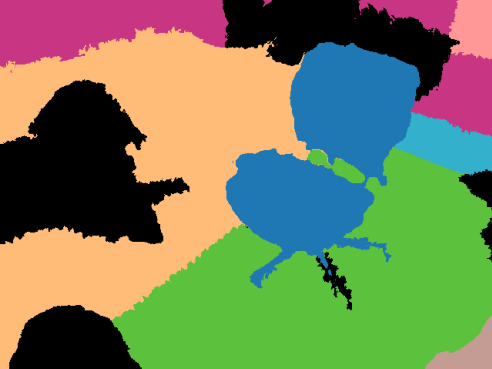}} 
\\
\rotatebox[origin=c]{90}{GT (RGB)} & 
\raisebox{-0.5\height}{\includegraphics[width=\sz\linewidth]{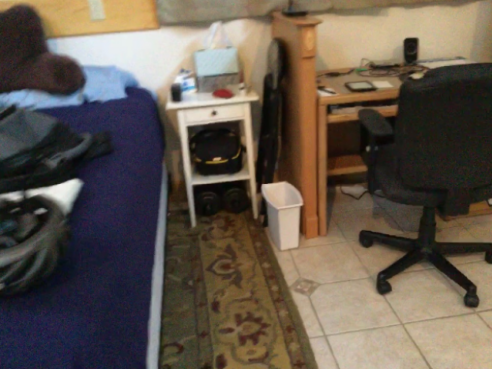}} 
& 
\raisebox{-0.5\height}{\includegraphics[width=\sz\linewidth]{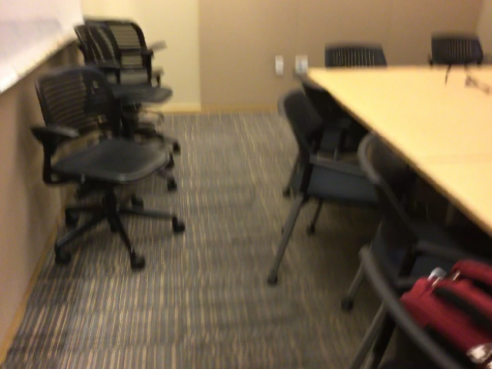}} 
&
\raisebox{-0.5\height}{\includegraphics[width=\sz\linewidth]{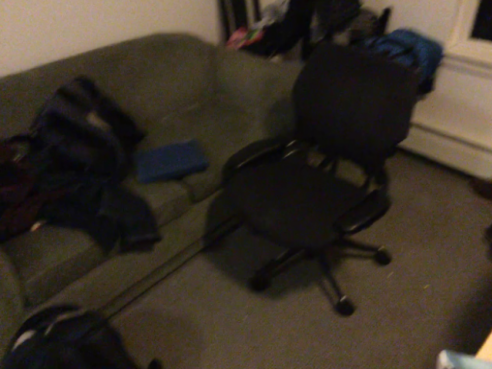}} 
\\
&\texttt{scene0000} & \texttt{scene0169} & \texttt{scene0207} 
\\

\end{tabular}
}
\vskip -0.08in
\caption{\textbf{Semantic Renderings on ScanNet}. Qualitative comparison on semantic synthesis of our method and baseline semantic SLAM method SGS-SLAM. Black areas in GT labels denote regions that are unannotated.
}
\label{fig:scannet_sem}
\vskip -0.15in
\end{figure}

As illustrated in Fig.~\ref{fig:scannet_sem}, our approach achieves superior semantic segmentation accuracy compared to the SGS-SLAM and GS$^3$LAM baselines. For example, in the first column of Fig.\ref{fig:scannet_sem}, our semantic rendering provides a more accurate representation of the desks, chairs, and night tables than baselines.

\section{Zero-Shot Results on NYUv2}
\begin{figure*}[htb]
\centering
{\footnotesize
\setlength{\tabcolsep}{1pt}
\renewcommand{\arraystretch}{0.9}
\newcommand{\sz}{0.15}
\begin{tabular}{ccccccc}
\rotatebox[origin=c]{90}{SplaTAM} & 
\raisebox{-0.5\height}{\includegraphics[width=\sz\linewidth]{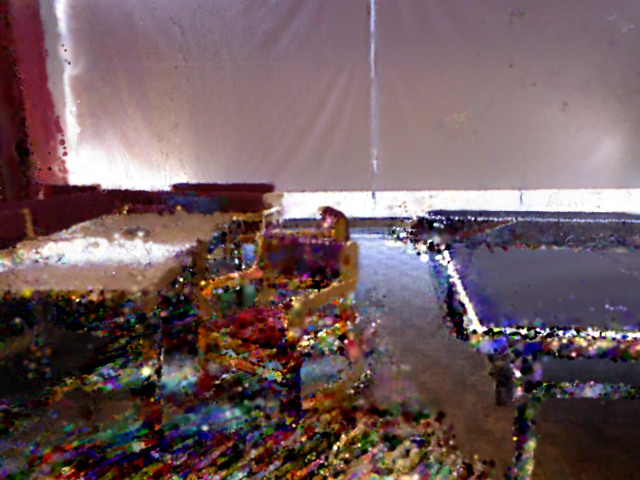}} 
& 
\raisebox{-0.5\height}{\includegraphics[width=\sz\linewidth]{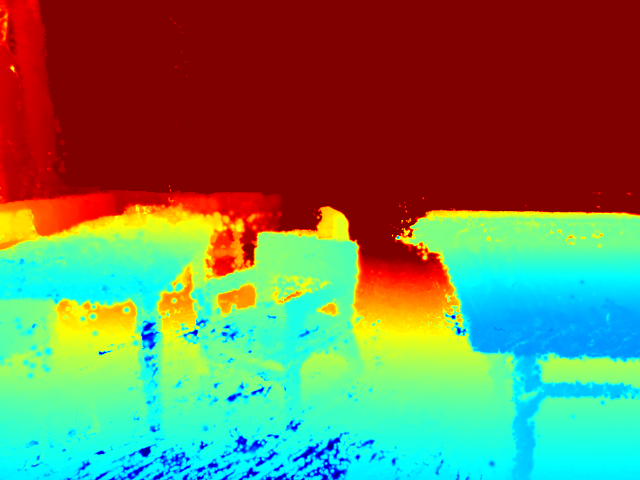}} 
&
\raisebox{-0.5\height}{\includegraphics[width=\sz\linewidth]{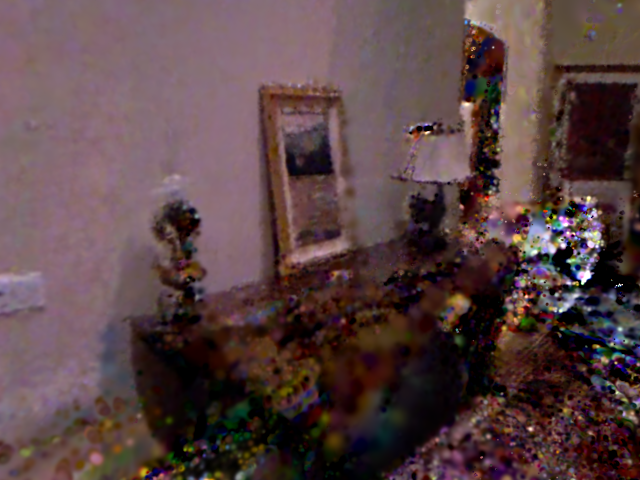} }
& 
\raisebox{-0.5\height}{\includegraphics[width=\sz\linewidth]{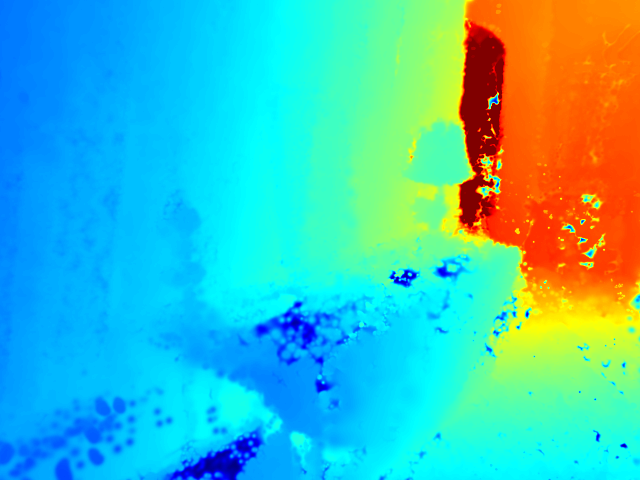}} 
&
\raisebox{-0.5\height}{\includegraphics[width=\sz\linewidth]{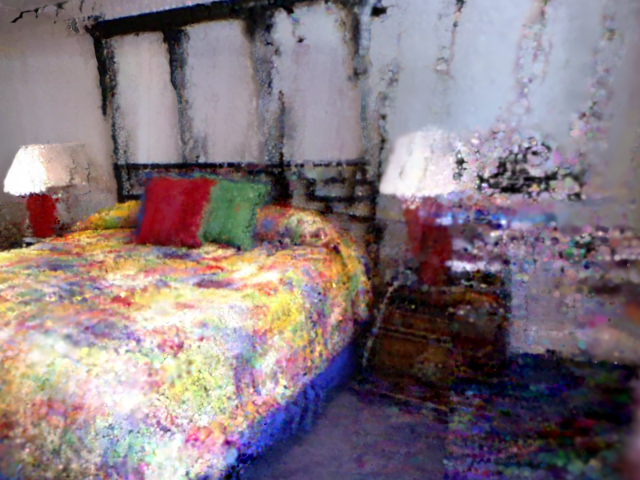} }
& 
\raisebox{-0.5\height}{\includegraphics[width=\sz\linewidth]{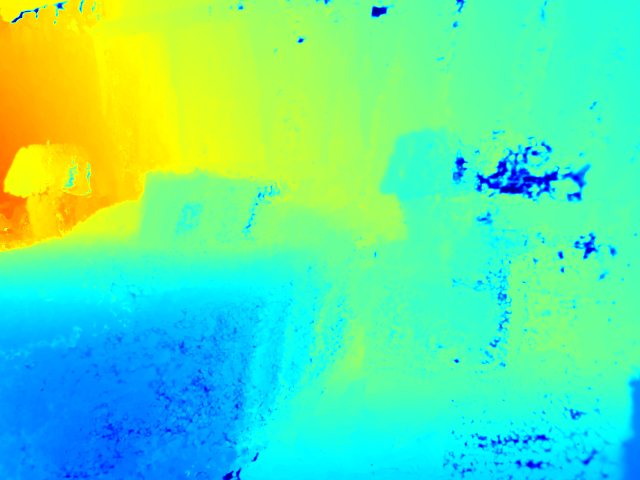}} 
\\
\rotatebox[origin=c]{90}{SGS SLAM} & 
\raisebox{-0.5\height}{\includegraphics[width=\sz\linewidth]{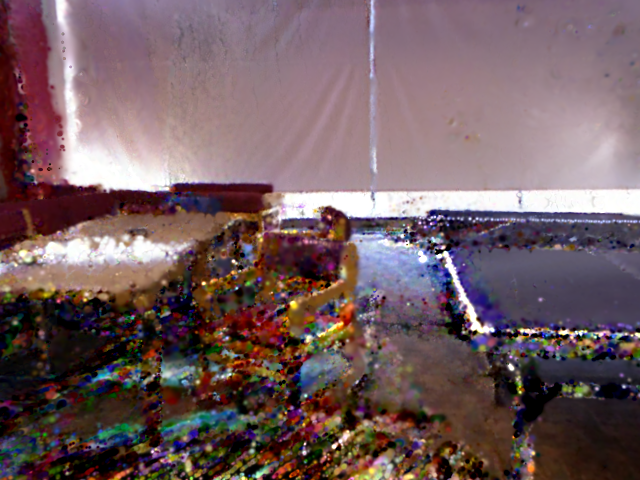}} 
& 
\raisebox{-0.5\height}{\includegraphics[width=\sz\linewidth]{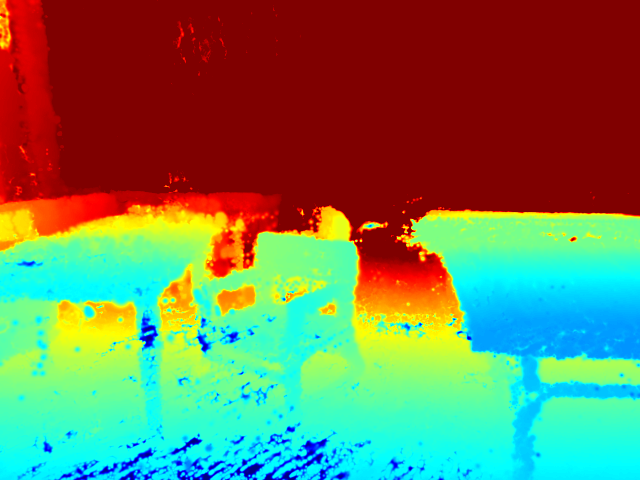}} 
&
\raisebox{-0.5\height}{\includegraphics[width=\sz\linewidth]{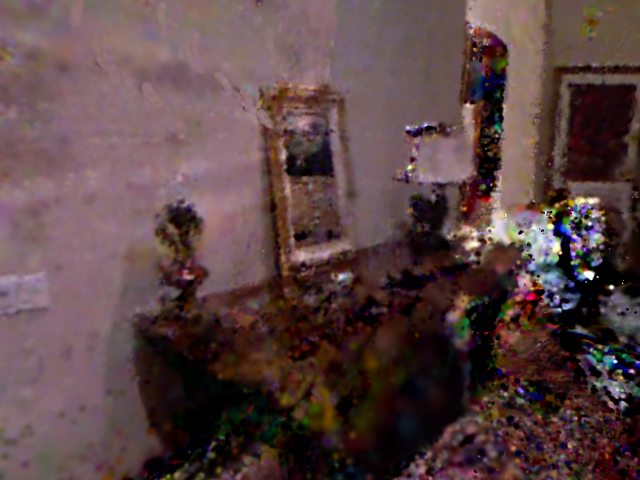}} 
& 
\raisebox{-0.5\height}{\includegraphics[width=\sz\linewidth]{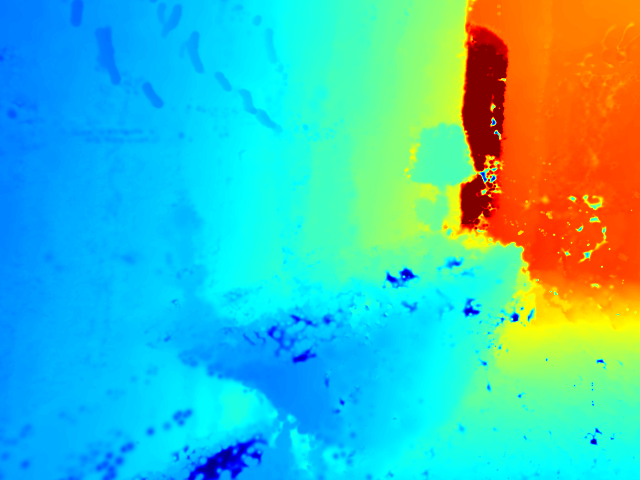}} 
&
\raisebox{-0.5\height}{\includegraphics[width=\sz\linewidth]{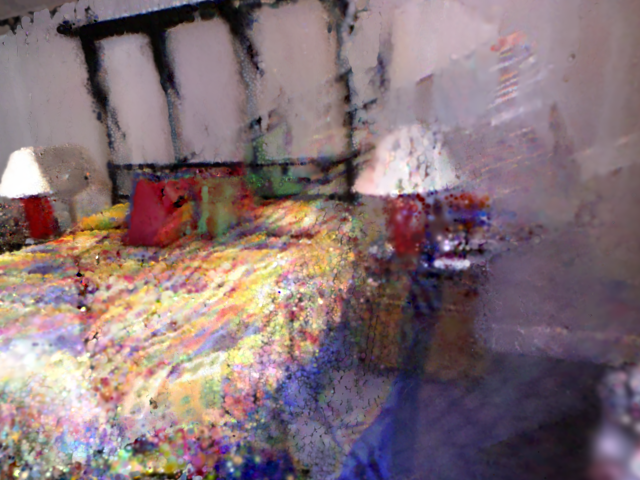}} 
& 
\raisebox{-0.5\height}{\includegraphics[width=\sz\linewidth]{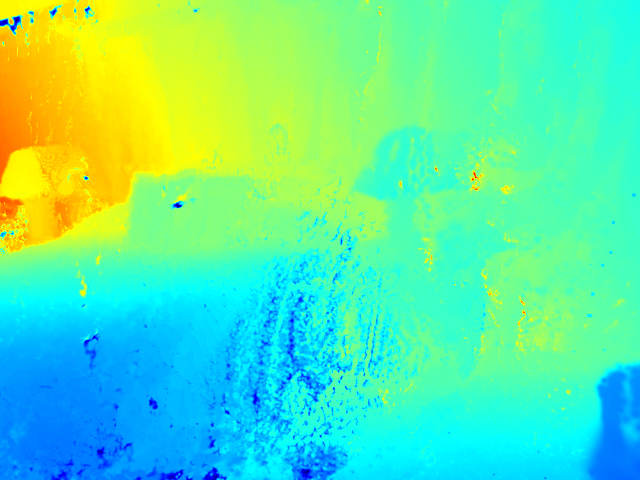}} 

\\
\rotatebox[origin=c]{90}{\textbf{GS4 (Ours)}} & 
\raisebox{-0.5\height}{\includegraphics[width=\sz\linewidth]{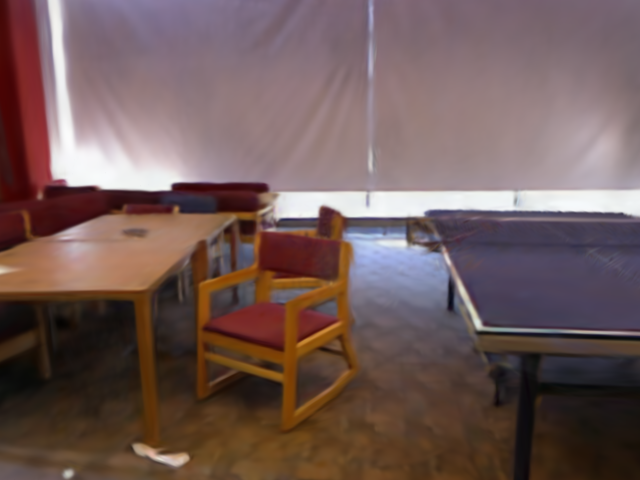}} 
& 
\raisebox{-0.5\height}{\includegraphics[width=\sz\linewidth]{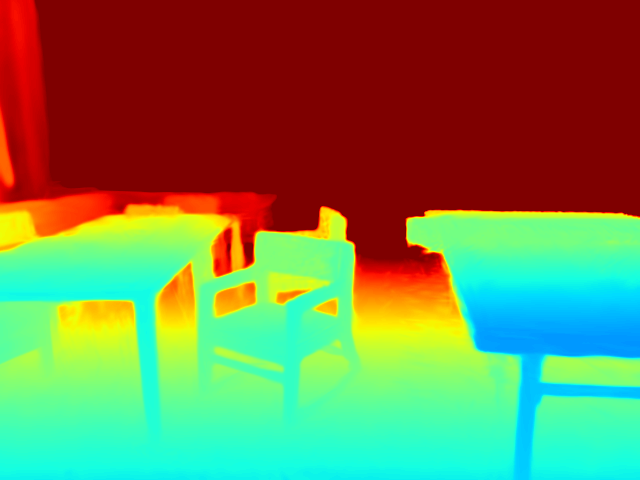}} 
&
\raisebox{-0.5\height}{\includegraphics[width=\sz\linewidth]{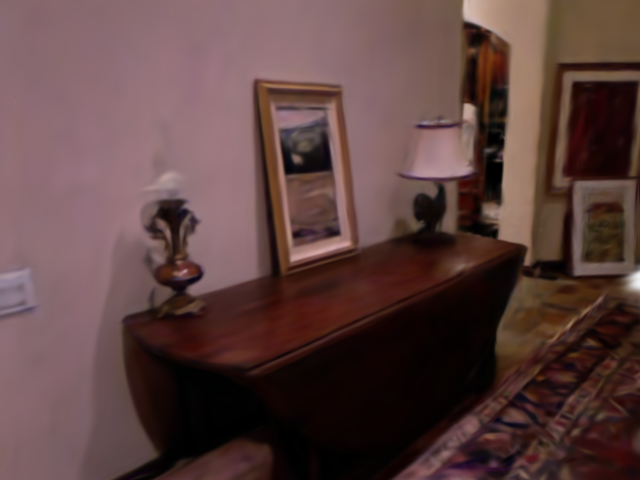}} 
&
\raisebox{-0.5\height}{\includegraphics[width=\sz\linewidth]{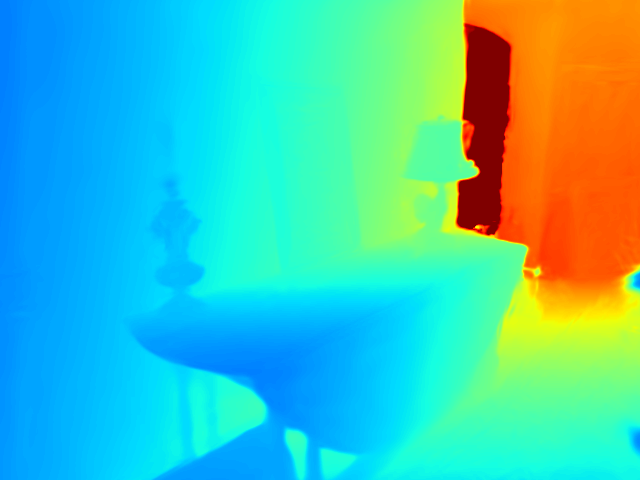}} 
&
\raisebox{-0.5\height}{\includegraphics[width=\sz\linewidth]{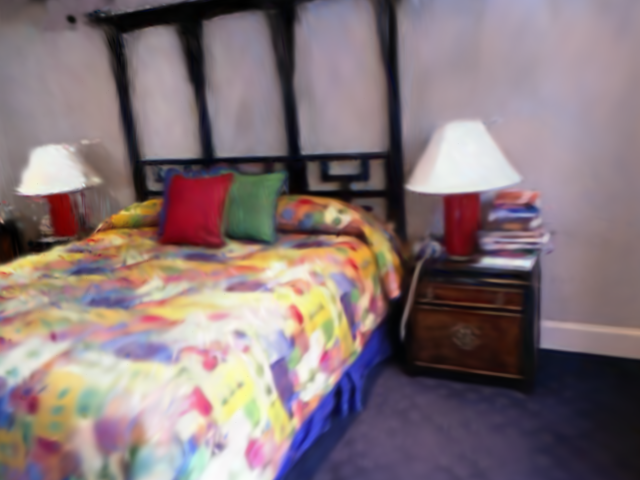}} 
&
\raisebox{-0.5\height}{\includegraphics[width=\sz\linewidth]{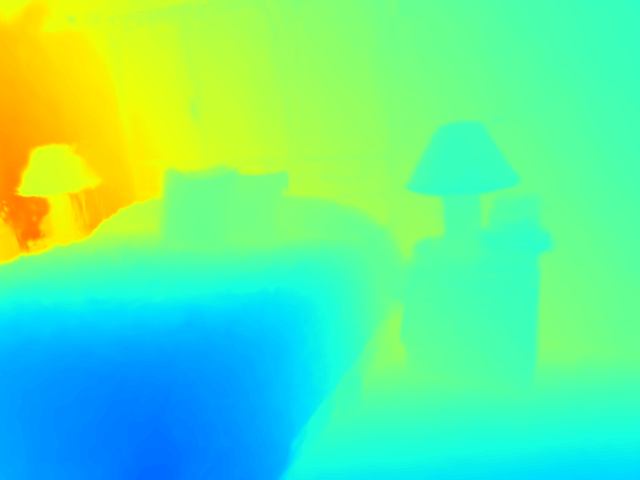}} 
\\
\rotatebox[origin=c]{90}{GT} & 
\raisebox{-0.5\height}{\includegraphics[width=\sz\linewidth]{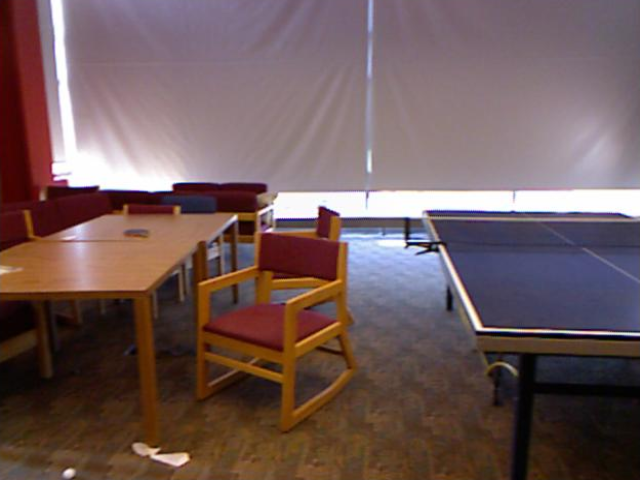}} 
& 
\raisebox{-0.5\height}{\includegraphics[width=\sz\linewidth]{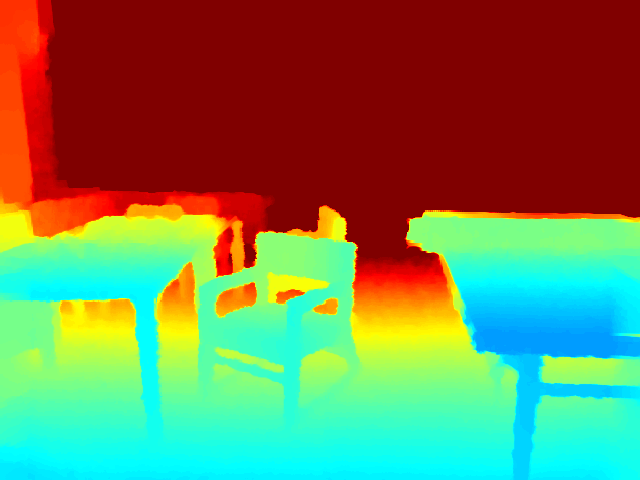}} 
&
\raisebox{-0.5\height}{\includegraphics[width=\sz\linewidth]{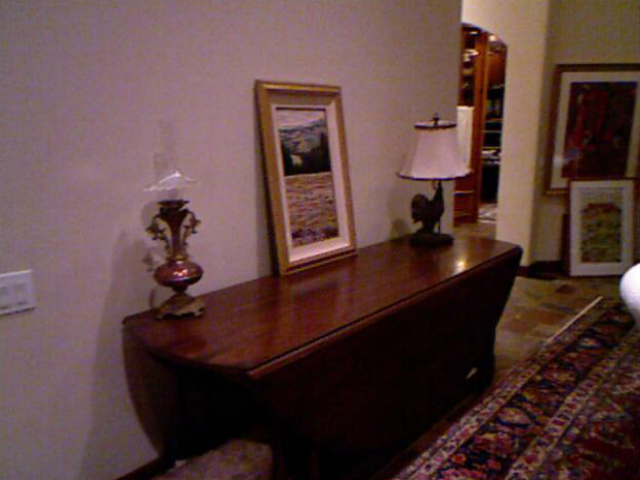}} 
& 
\raisebox{-0.5\height}{\includegraphics[width=\sz\linewidth]{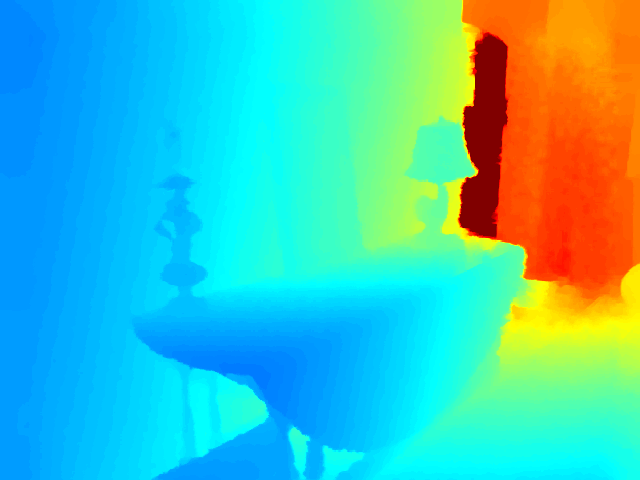}} 
&
\raisebox{-0.5\height}{\includegraphics[width=\sz\linewidth]{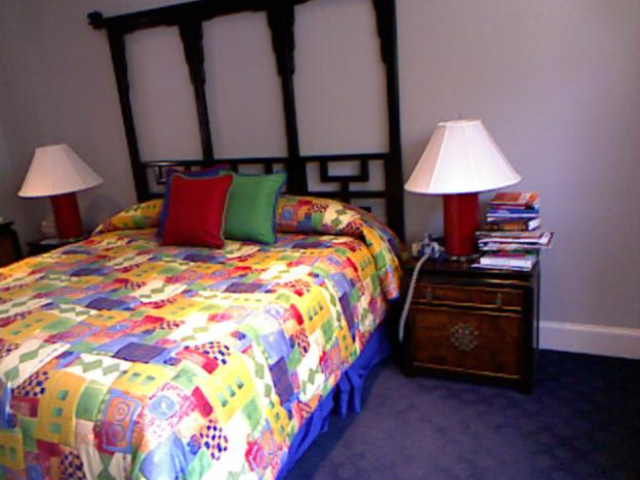}} 
& 
\raisebox{-0.5\height}{\includegraphics[width=\sz\linewidth]{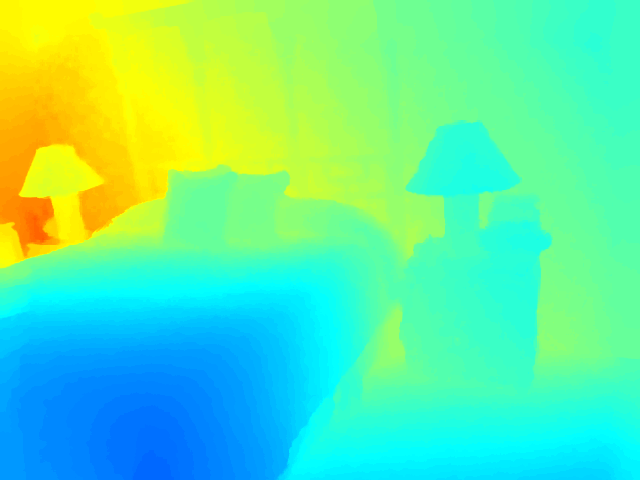}} 
\\
& \multicolumn{2}{c}{\texttt{sudent lounge}} & \multicolumn{2}{c}{\texttt{dining room}} & \multicolumn{2}{c}{\texttt{bedroom}} 
\\

\end{tabular}
}
\vskip -0.05in
\caption{\textbf{Zero-shot Visualization on NYUv2}. Qualitative comparison of our method and other GS-based SLAM methods. 
}
\vskip -0.1in
\label{fig:nyuv2_vis}
\end{figure*}

Fig.~\ref{fig:nyuv2_vis} illustrates our zero-shot visualization results on the NYUv2 dataset. Despite our models being exclusively trained on the ScanNet dataset, our method demonstrates superior performance on the NYUv2 dataset compared to other GS-based SLAM approaches. In Table~\ref{tab:nyutable_scene}, we present the quantitative zero-shot results across three scenes from the NYUv2 dataset. Our method outperforms all other GS-based SLAM approaches on all rendering metrics, while using significantly fewer Gaussians.

\begin{table}[htb]
\begin{center}
\caption{Rendering and Runtime performance on NYUv2 test scenes with $640 \times 480$ input.  
% Best results are highlighted as \colorbox{colorFst}{\bf first}, \colorbox{colorSnd}{second}. 
GS Num represents the number of 3D Gaussians included in the scene after mapping is complete. FPS is conducted on an Nvidia RTX TITAN.}
\label{tab:nyutable_scene}
\resizebox{0.99\linewidth}{!}{%
  \begin{small}
\begin{tabular}{llcccc}
\toprule
\bf Methods  & \bf Metrics & bedroom & student lounge & dining room & \bf Avg \\
\midrule
&PSNR$\uparrow$ & 17.99 & \nd 20.77 & 17.82 & 18.86 \\
 & SSIM$\uparrow$ & 0.692 & \nd 0.795 & 0.589 & 0.692 \\
SplaTAM &LPIPS$\downarrow$ & 0.343 & \nd 0.309 & 0.465 & 0.372 \\
    &GS Num$\downarrow$ & 1529k & 1116k & 1063k & 1236k \\
    % &FPS$\uparrow$ & 0.32 & 0.33 & 0.33 & 0.33 \\
\cdashmidrule{1-6}
&PSNR$\uparrow$ & 10.81 & 12.94 & 11.76 & 11.84 \\
 & SSIM$\uparrow$ & 0.146 & 0.299 & 0.217 & 0.221 \\
RTG-SLAM &LPIPS$\downarrow$ & 0.738 & 0.662 & 0.709 & 0.703 \\
    &GS Num$\downarrow$ & \nd 906k & \nd 591k & \nd 925k & \nd 807k \\
    % &FPS$\uparrow$ & \nd 1.16 & \nd 1.3 & \nd 1.15 & \nd 1.20 \\
\cdashmidrule{1-6}
&PSNR$\uparrow$ & \nd 19.66 & 20.41 & \nd 17.90 & \nd 19.32 \\
 & SSIM$\uparrow$ & \nd 0.754 & 0.780 & \nd 0.590 & \nd 0.708 \\
SGS-SLAM &LPIPS$\downarrow$ & \nd 0.289 & 0.318 & \nd 0.463 & \nd 0.357 \\
    &GS Num$\downarrow$ & 1201k & 1074k & 1049k & 1108k \\
    % &FPS$\uparrow$ & 0.25 & 0.25 & 0.24 & 0.25 \\
\cdashmidrule{1-6}
     &PSNR$\uparrow$ & \fs 21.15 & \fs 22.35 & \fs 23.21 &  \fs 22.24 \\
 & SSIM$\uparrow$ &  \fs 0.885 &  \fs 0.867 &  \fs 0.846 &  \fs 0.866 \\
\textbf{GS4 (Ours, 2nd stg)} &LPIPS$\downarrow$ & \fs  0.217 &  \fs 0.238 &  \fs 0.307 &  \fs 0.254 \\
    &GS Num$\downarrow$ &  \fs 273k & \fs 228k &  \fs 392k &  \fs 298k \\
    % &FPS$\uparrow$ &  &  &  &  \\
\bottomrule
\end{tabular}
\end{small}
  }
\end{center}\vskip -0.15in
\vskip -0.1in
\end{table}

\section{Zero-Shot Results on TUM RGB-D}

In Table~\ref{tab:tumtable_scene}, we present the quantitative zero-shot results across three scenes from the TUM RGB-D dataset. TUM RGB-D provides ground truth camera trajectories, so we also report tracking performance. Our method achieves rendering performance comparable to that of all other GS-based SLAM approaches, while using significantly fewer Gaussians.
\begin{table}[htb]
\begin{center}
\caption{Rendering, Tracking, and Runtime performance on TUM RGB-D test scenes with $640 \times 480$ input.  
% Best results are highlighted as \colorbox{colorFst}{\bf first}, \colorbox{colorSnd}{second}. 
GS Num represents the number of 3D Gaussians included in the scene after mapping is complete. FPS is conducted on an Nvidia RTX TITAN.}
\label{tab:tumtable_scene}
\resizebox{0.99\linewidth}{!}{%
  \begin{small}
\begin{tabular}{llcccc}
\toprule
\bf Methods  & \bf Metrics & fr1$\_$desk & fr2$\_$xyz & fr3$\_$office & \bf Avg \\
\midrule
&PSNR$\uparrow$ & \nd 22.07 & \nd 24.66 & \nd 21.54 & \fs 22.76 \\
 & SSIM$\uparrow$ & 0.857 & \fs 0.947 & \nd 0.870 & \nd 0.891 \\
SplaTAM &LPIPS$\downarrow$ & \nd 0.238 & \nd 0.099 & \fs 0.210 & \fs 0.182 \\
&ATE RMSE$\downarrow$ & 3.33 & 1.55 & 5.28 & 3.39 \\
    &GS Num$\downarrow$ & 969k & 635k & 806k & 803k \\
    % &FPS$\uparrow$ & 0.32 & 0.05 & 0.25 & 0.20 \\
\cdashmidrule{1-6}
&PSNR$\uparrow$ & 18.49 & 20.18 & 20.59 & 19.75 \\
 & SSIM$\uparrow$ & 0.715 & 0.795 & 0.797 & 0.769 \\
RTG-SLAM &LPIPS$\downarrow$ & 0.438 & 0.353 &  0.394 & 0.395 \\
&ATE RMSE$\downarrow$ & \fs 1.66 & \fs 0.38 & \fs 1.13 & \fs 1.06 \\
    &GS Num$\downarrow$ & 236k & \fs 84k & 273k & 198k \\
    % &FPS$\uparrow$ & \fs 2.07 & 2.12 & \fs 1.97 & 2.05 \\
\cdashmidrule{1-6}
&PSNR$\uparrow$ & \fs 22.10 & \fs 25.61 & 19.62 & \nd 22.44 \\
 & SSIM$\uparrow$ & \fs 0.886 & \nd 0.946 & 0.796 & \nd 0.876 \\
SGS-SLAM &LPIPS$\downarrow$ & \fs 0.176 & \fs 0.097 & 0.280 & \nd 0.184 \\
&ATE RMSE$\downarrow$ & 3.57 & 1.29 & 9.08 & 4.65 \\
    &GS Num$\downarrow$ & 808k & 695k & 701k & 735k \\
    % &FPS$\uparrow$ & 0.25 & 0.10 & 0.20 & 0.18 \\
\cdashmidrule{1-6}
     &PSNR$\uparrow$ & 21.71 & 23.86 & \fs 22.54 & \nd 22.70 \\
 & SSIM$\uparrow$ & \nd 0.877 & 0.904 & \fs 0.890 & \fs 0.903 \\
\textbf{GS4 (Ours, 2nd stg)} &LPIPS$\downarrow$ & 0.242 &  0.154 & \nd 0.226 & 0.191 \\
      &ATE RMSE$\downarrow$ & \nd 1.86 & \nd 0.63  & 1.95 & 1.48 \\
    &GS Num$\downarrow$ & \fs 175k & \nd 87k & \fs 190k & \fs 166k\\
    % &FPS$\uparrow$ & \nd  & \fs 3.7 & 1.64  \\
\bottomrule
\end{tabular}
\end{small}
  }
\end{center}
\vskip -0.1in
\end{table}

% {
%     \small
%     \bibliographystyle{ieeenat_fullname}
%     \bibliography{main}
% }

% \end{document}